%% file: Main.tex
\documentclass[Afour,sageh,times]{sagej}

\usepackage{moreverb,url}

\usepackage[colorlinks,bookmarksopen,bookmarksnumbered,citecolor=red,urlcolor=red]{hyperref}

\usepackage{calligra}
\usepackage{subcaption}
\usepackage{booktabs}
\usepackage{amsmath}
\usepackage{breqn}
\usepackage{sidecap}
\usepackage{balance}
\usepackage{textcomp}
\usepackage{float}
\usepackage{graphicx}
\usepackage{amsfonts}
\usepackage{amssymb}
\usepackage{eucal}
\usepackage{siunitx}

\newsavebox{\twosubbox}
\usepackage{url}
\usepackage{bm}
\usepackage[font=footnotesize, skip=5pt]{caption}
\usepackage[normalem]{ulem} 

\newcommand\BibTeX{{\rmfamily B\kern-.05em \textsc{i\kern-.025em b}\kern-.08em
T\kern-.1667em\lower.7ex\hbox{E}\kern-.125emX}}
\newcommand{\x}{\bm{x}} 
\newcommand{\z}{\bm{z}} 

\newcommand{\Metric}{\bm{M}} 
\newcommand{\curve}{c} 
\newcommand{\Jac}{\bm{J}} 
\newcommand{\trsp}{\mathsf{T}} 

\newcommand{\ambient}{\mathcal{X}} 
\newcommand{\latent}{\mathcal{Z}} 
\newcommand{\Manifold}{\mathcal{M}} 
\newcommand{\R}{\mathbb{R}} 
\newcommand{\Sph}{\mathcal{S}} 
\newcommand{\Q}{\mathcal{Q}} 

\newcommand{\I}{\mathbb{I}} 
\newcommand{\Prob}{p} 
\newcommand{\vMF}{\mathrm{vMF}}
\newcommand{\Length}{\mathcal{L}} 
\newcommand{\q}{\bm{q}} 
\newcommand{\Loss}{\mathcal{L}} 

\newcommand{\g}{\bm{g}} 

\usepackage{xcolor}
\definecolor{green_new}{HTML}{008000}

\newcommand{\etal}{\MakeLowercase{\textit{et al.}}}

\def\volumeyear{2021}
\begin{document}

\runninghead{Beik-Mohammadi \etal}

\title{Reactive Motion Generation on Learned Riemannian Manifolds}

\author{Hadi Beik-Mohammadi\affilnum{1}\affilnum{2}, S\o{}ren Hauberg\affilnum{3}, Georgios Arvanitidis\affilnum{3}, Gerhard Neumann\affilnum{2}, and Leonel Rozo\affilnum{1} }

\affiliation{\affilnum{1} Bosch Center for Artificial Intelligence (BCAI), Renningen, Germany\\
\affilnum{2} Autonomous Learning Robots Lab, Karlsruhe Institute of Technology (KIT), Karlsruhe, Germany\\
\affilnum{3} Section for Cognitive Systems, Technical University of Denmark (DTU), Lyngby, Denmark}

\corrauth{Hadi Beik-Mohammadi, Bosch Center for Artificial Intelligence (BCAI)
Reinforcement learning and planning (CR/PJ-AI-R31),
70049 Stuttgart,
GERMANY}

\email{hadi.beik-mohammadi@de.bosch.com}

\begin{abstract}
In recent decades, advancements in motion learning have enabled robots to acquire new skills and adapt to unseen conditions in both structured and unstructured environments. In practice, motion learning methods capture relevant patterns and adjust them to new conditions such as dynamic obstacle avoidance or variable targets. In this paper, we investigate the robot motion learning paradigm from a Riemannian manifold perspective. We argue that Riemannian manifolds may be learned via human demonstrations in which geodesics are natural motion skills. The geodesics are generated using a learned Riemannian metric produced by our novel variational autoencoder (VAE), which is intended to recover full-pose end-effector states and joint space configurations. In addition, we propose a technique for facilitating on-the-fly end-effector/multiple-limb obstacle avoidance by reshaping the learned manifold using an obstacle-aware ambient metric. The motion generated using these geodesics may naturally result in multiple-solution tasks that have not been explicitly demonstrated previously. We extensively tested our approach in task space and joint space scenarios using a $7$-DoF robotic manipulator. We demonstrate that our method is capable of learning and generating motion skills based on complicated motion patterns demonstrated by a human operator. Additionally, we assess several obstacle avoidance strategies and generate trajectories in multiple-mode settings.
\end{abstract}


\maketitle

\section{Introduction}
\input{Sections/Introduction}

\section{Background and Related Work}
\input{Sections/Related_Work}

\input{Sections/Approach}

\section{Experiments} 
\input{Sections/Experiments}

\section{Discussion} 
\input{Sections/Discussion}


%




\begin{acks}
This work was supported by a research grant (42062) from \textsc{villum fonden}. This project has also received funding from the European Research Council (\textsc{erc}) under the European Union’s Horizon 2020 research and innovation programme (grant agreement 757360). Our work was funded in part by the Novo Nordisk Foundation through the Center for Basic Machine Learning Research in Life Sciences (\textsc{mlss}) with reference number NNF20OC0062606.
\end{acks}




\bibliographystyle{SageH}
\bibliography{references.bib}

\end{document}

%% file: Sections/Introduction.tex
    Robot learning has gained interest recently because of its potential to endow robots with a repertoire of motion skills to operate autonomously or assist humans in repetitive, precise, and risk-averse tasks. 
    In this paper, we tackle the robot motion generation problem from a robot learning perspective, which involves modeling and reproducing motion skills such as picking, placing, etc. 
    In practice, we may broadly classify motion generation techniques into two categories: Classic motion planners and movement primitives.
    The former generates obstacle-free continuous trajectories between start and goal configurations using search or sampling-based algorithms~\citep{Elbanhawi14:MotionPlanning,Mohanan18:DynamicMotPlan}.
    Movement primitives, in contrast, are a modular abstraction of robot movements, where a primitive represents an ``atomic action" or an ``elementary movement", which are often designed using robot learning techniques~\citep{Schaal03}.  
    A relevant distinction between the aforementioned approaches is that motion planners require a precise description of the robot workspace to generate a continuous path to the goal, while movement primitives rely on observed motion patterns to encapsulate the desired trajectory without an explicit model of the environment. 
    If an unseen obstacle shows up in the robot workspace, motion planners often need to replan the entire trajectory from scratch, while most movement primitives require to be retrained, unless they are provided with via-point passing features.
    We aim for a trade-off between these two categories: A learning-based reactive motion generation that generates robot movements resembling the observed motion patterns while avoiding obstacles on the fly.
    
    Human movements often involve following geodesic paths, which are the most efficient and natural-looking paths. This is particularly true when completing tasks that require hand skills, such as reaching for an object. In these cases, individuals tend to take the path that minimizes the distance and energy required for the movement. This has been observed in different studies~\citep{GeodesicHumanMovement12Moon, GeodesicHumanMovement22Klein}. Therefore, 
    we resort to learning-from-demonstration (LfD) to build a model of robot motions, a technique in which the skill model is learned by encapsulating motion patterns from human demonstrations~\citep{Osa2018:Imitation}. 
    The current LfD approaches do not need an environment model and some can quickly adapt to settings with dynamic targets~\citep{Osa2018:Imitation, Ijspeert13:dmp, calinon2016:tutorial}. 
    We can broadly identify three main groups of LfD approaches that use motion primitives as their building blocks. 
    The first group explicitly considers the motion dynamics~\citep{Ijspeert13:dmp, MixtureOfAttractors2018:Manschitz}. 
    The second group builds on probabilistic approaches that exploit demonstration variability for motion adaptation and control~\citep{Huang19:KMP, calinon2016:tutorial, paraschos2018:ProMP}. 
    The last group uses neural networks as their skill model and focuses on robot motion generalization~\citep{Yunus2019:CNMPs, GoalConditionVAE2019:Osa}.
    Despite their significant contributions (see Section~\ref{sec:background}), several challenges are still open: Encoding and reproducing full-pose end-effector movements, skill adaptation to unseen or dynamic obstacles at both task and joint spaces, handling multiple-solution (a.k.a. multiple-mode) tasks in the aforementioned spaces, and generalization to unseen situations (e.g. new target poses), among others.

    In our previous paper~\citep{GeodesicMotionSkill2021:BeikMohammadi}, we provided an LfD approach that addressed several of the foregoing problems by leveraging a Riemannian perspective for learning robot motions using task space demonstrations.
    We employed a Riemannian formulation to represent a motion skill, in which human demonstrations were assumed to form a Riemannian manifold (i.e. a smooth surface), which could be learned in task space $\mathbb{R}^3 \times \mathcal{S}^3$. 
    This approach showed to have a number of advantages: Strong capabilities to model the non-linearity of motion data, adaptability to unseen conditions (e.g. obstacles), and generation of novel task solutions.
    The core of our previous approach was built on a variational autoencoder (VAE) that was exploited to compute a Riemannian metric describing the underlying structure of these skill manifolds. 
    This metric allowed us to measure distances on Riemannian manifolds, a property that is used to compute geodesics (i.e., the shortest paths on Riemannian manifolds), which was then leveraged as our motion generation mechanism.
    In addition, we designed an ambient space metric to reshape the skill manifold on the fly in order to perform obstacle avoidance. 
    Furthermore, the learned skill manifold was able to encode multiple-solution tasks, which naturally resulted in novel hybrid solutions based on the synergy of a subset of the demonstrations.
    
    \textbf{In this paper}, we extend our previous work in two different ways. 
    First, inspired by the need of avoiding obstacles at any location of the robot body, we propose a new reactive motion generation method that also leverages the Riemannian approach proposed in~\citep{GeodesicMotionSkill2021:BeikMohammadi} for joint space skills.
    To do so, we develop a new VAE architecture that integrates the robot's forward kinematics to access task space information at any point on the robot body. 
    This new approach makes joint space skill adaptation possible, allowing the robot to simultaneously avoid unseen and dynamic obstacles and handle multiple-solution (a.k.a. multiple-mode) tasks in both task and joint spaces. 
    Second, we describe both methodologies with simple examples and extensively evaluate them on real tasks using a $7$-DoF robot manipulator. 
    The experiments showcase how our techniques allow a robot to learn and reconstruct motion skills of varying complexity.
    In particular, we show how our methods enable obstacle avoidance at task and joint space levels. 
    Moreover, we provide experiments where the robot exploits multiple-solution tasks to effectively generate hybrid solutions without model retraining. 
    Furthermore, we experimentally analyze how the latent space dimensionality impacts the quality of the geodesic motion generator, and how the robot behaves when controlled by geodesics crossing the manifold boundaries.
    
    In summary, we propose a new Riemannian learning framework for reactive motion skills that lies on the middle ground between classical motion planning and motion primitives. 
    Similar to motion primitives, we exploit human demonstrations to learn a model of a skill with the assumption that these demonstrations belong to a smooth surface characterized as a Riemannian manifold. 
    Additionally, our method, like motion planners, derives a time-independent reference trajectory from a geodesic motion generator while incorporating obstacle avoidance capabilities. 
    
    In Section~\ref{sec:background}, we review relevant work on learning from demonstration, variational auto-encoders, Riemannian manifolds, their connections, and ambient space metrics. 
    In Section~\ref{sec:riemannian_manifolds}, we examine how a skill manifold may be trained and used to reconstruct robot motion when the demonstrations are provided in the task and joint spaces. 
    Section~\ref{sec:geodesic_motion} investigates geodesic computation and their ability to avoid unseen obstacles on the fly.
    In Section~\ref{sec:experiments}, we show our method capabilities via a series of real-world robot experiments.
    Finally, in Section~\ref{sec:discussion}, we discuss some of the limitations of our approach and provide possible solutions to be explored in future work.

%% file: Sections/Related_Work.tex
\label{sec:background}

\begin{figure}[t]
\centering
\begin{subfigure}{\linewidth}
   \includegraphics[width=1\linewidth]{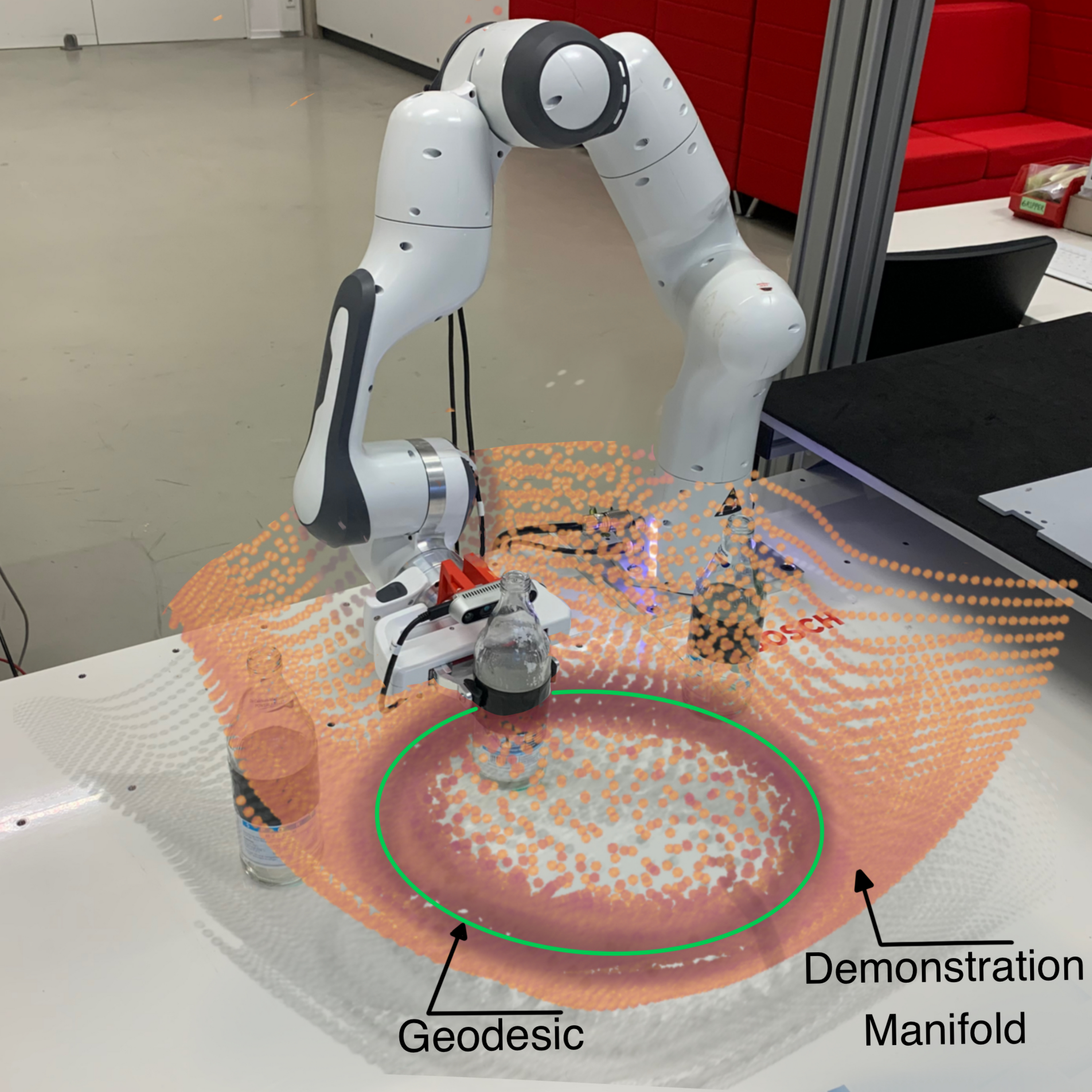}
\end{subfigure}
\caption{From demonstrations we learn a variational autoencoder that spans a random Riemannian manifold. 
Geodesics on this manifold are viewed as motion skills. }
\label{fig:Teaser}
\end{figure}

Due to the strong connections between our work and the learning from demonstration (LfD) technique, we begin this section by discussing pertinent work on learning and synthesizing joint and task space motion skills via LfD. 
We look into how these approaches reconstruct full-pose trajectories, provide obstacle avoidance capabilities, and reproduce multiple-solution (multiple-mode) tasks. 
Second, we review Riemannian geometry as well as variational auto-encoders (VAEs), which are essential concepts for our approach. 
Finally, we explain recent advances in the connection between VAEs and Riemannian geometry, which is the backbone of our methods.

\subsection{Learning from Demonstration (LfD)}
    LfD is a robot programming technique that leverages human demonstrations, recorded via kinesthetic teaching or teleoperation, to learn a model of a task~\citep{Ravichandar20:LfD}. 
    End-effector positions and orientations, joint configurations, linear or angular velocities, and accelerations are all examples of data that can be used in LfD.
    The methods for dealing with motion dynamics are outside the scope of this study; Instead, we focus on techniques for learning and synthesizing robot skills built on joint and task space trajectories.
    
    We identify three key lines of research and their particular features (e.g. obstacle avoidance) for robot motion learning. 
    The first group focuses on motion dynamics learning~\citep{Ijspeert13:dmp, MixtureOfAttractors2018:Manschitz}, where demonstrations are considered solutions to specific dynamical systems. 
    These techniques are well-behaved when confronted with changes in the environment due to their autonomous systems formulation. 
    The second set of approaches builds on probabilistic methods that exploit data variability and model uncertainty~\citep{Huang19:KMP, calinon2016:tutorial, paraschos2018:ProMP}. 
    Their probabilistic formulation allows robots to execute the skill using a large diversity of trajectories sampled from the skill model.
    The last category includes approaches that use neural networks for increasing generalization in robot motion learning~\citep{Yunus2019:CNMPs, GoalConditionVAE2019:Osa}. 
    Furthermore, there exist methods that combine dynamical systems and neural networks~\citep{Bahl20:NDPs, EuclidFlow2020:Rana}, or dynamical systems and probabilistic models~\citep{Ugur2020:CompliantDMP, Khansari2011:StabNonDynSys}. 
    All of the aforementioned methods belong to the category of movement primitives (MPs), which is an alternative approach to classic motion planning~\citep{Elbanhawi14:MotionPlanning} for robot motion generation. 

    In contrast, we propose a reactive motion generation technique that lies on a middle ground between movement primitives and motion planning. 
    Similar to motion primitives, we exploit human demonstrations to learn a model of a skill with the assumption that these demonstrations belong to a smooth surface characterized as a Riemannian manifold. 
    Additionally, our method, like motion planners, derives a time-independent reference trajectory generated from geodesic curves, which can be locally deformed to avoid unseen obstacles.
    Specifically, our method leverages a neural network (VAE) to learn a Riemannian metric that incorporates the network uncertainty. 
    This metric allows us to generate motions that resemble the demonstrations via geodesics.
    The decoded geodesics are then used as reference trajectories on the robot to reproduce motion trajectories that resemble the demonstrations. 

    Complex robot movements may involve sophisticated full-pose end-effector trajectories, making it imperative to have a learning framework capable of encoding full-pose motion patterns.
    The main challenge is then how to properly encode and reproduce orientation movements. 
    Despite most LfD approaches have overlooked this problem~\citep{paraschos2018:ProMP, Yunus2019:CNMPs, calinon2016:tutorial, Huang19:KMP}, recent works have addressed it using probabilistic models~\citep{OrientationProMP2021:Rozo, Zeestraten17:riemannian}.
    In our previous work~\citep{GeodesicMotionSkill2021:BeikMohammadi}, we proposed a VAE architecture capable of encoding full-pose trajectories, which is here exploited for learning a variety of real robotic tasks.
    
    Obstacle avoidance is another feature that a reactive motion generation should offer. 
    While several approaches rely solely on obstacles information given before learning~\citep{ObstacleAvoidanceRL1992:Prescott, VisionReactivePolicies2020:Aljalbout}, these are ineffective in dynamic environments. 
    Other techniques exploit via-points~\citep{paraschos2018:ProMP, Yunus2019:CNMPs, Huang19:KMP} to tackle this problem in an indirect fashion, but they do not require retraining the skill model. 
    A different perspective on the obstacle avoidance problem is to see obstacles as costs in an optimization framework that seeks to generate optimal and obstacle-free trajectories~\citep{Urainetal2021:ComposableEnergy, RMP}. 
    
    Meanwhile, our method tackles the obstacle avoidance problem from a metric-reshaping viewpoint. 
    We design obstacle-aware ambient space metrics to reshape the learned Riemannian metric. 
    The combination of these two metrics yields a new metric that is exploited to generate trajectories that simultaneously follow the demonstrations while also avoiding obstacles. 
    Note that the ambient space metric always represents a notion of distance to the obstacles in task space. 
    It is worth mentioning that the choice of demonstration ambient space defines how this metric can be designed to avoid obstacles. 
    For example, when using joint space demonstrations, the ambient space metric can incorporate information about the distance from any point of the robot to the obstacles, resulting in a multiple-limb obstacle avoidance capability.
    On the contrary, when using task space end-effector demonstrations, this metric only provides obstacle avoidance at the level of the robot end-effector. 

    Our obstacle avoidance technique employs a technique similar to that of CHOMP~\citep{Ratliff2009:chomp}, which involves the use of a simplified geometric representation of the robot and obstacles in order to construct a uniform grid in the task space. This grid is used to determine the potential for collision between the trajectory of the robot and any obstacles present. Additionally, our method incorporates this information into the geodesic energy leading to optimal paths that adapt accordingly. This approach is conceptually comparable to CHOMP, which also utilizes a cost function that considers the presence of obstacles.

    Human demonstrations can naturally show various solutions to a single motion skill~\citep{Rozo11:multipleSol, Yunus2019:CNMPs}, which is typically addressed using hierarchical techniques~\citep{Konidaris12:SkillTrees, Ewertonetal2015:multcollab}. 
    Our approach enables the encoding of multiple-solution tasks on the learned Riemannian manifold, which is then exploited to replicate the demonstrated skill and generate novel hybrid solutions based on a synergy of a (sub)set of demonstrations. 
    These hybrid solutions naturally emerge from our geodesic motion generator on the learned Riemannian manifold. 
    Note that the aforementioned robot motion learning techniques only provide trajectories that strictly follow the demonstrated patterns without the ability to generate hybrid solutions.

\subsection{Variational auto-encoders (VAEs)}
    VAEs are generative models~\citep{kingma:autoencoding} that learn and reconstruct data by encapsulating their density into a lower-dimensional latent space $\latent$. 
    Specifically, VAEs encode the training data density $\Prob(\x)$ in an ambient space $\ambient$ through a low-dimensional latent variable $\z$. 
    For simplicity, we consider the generative process of a Gaussian VAE defined as,
    \begin{align}
        \Prob(\z) &= \mathcal{N}\left(\z | \bm{0}, \I_d\right), & \z \in \latent ; \\
        \Prob_{\bm{\phi}}(\x|\z) &= \mathcal{N}\left(\z|\mu_{\bm{\phi}}(\z), \I_D \sigma_{\bm{\phi}}^2(\z)\right), & \x \in \ambient.
        \label{eq:vae_gen}
    \end{align}
    where $\mu_{\bm{\phi}} : \latent \rightarrow \ambient$ and $\sigma_{\bm{\phi}} : \latent \rightarrow \R^D_+$ are deep neural networks with parameters $\bm{\phi}$ estimating the mean and the variance of the posterior distribution $\Prob_{\bm{\phi}}(\x|\z)$, and $\I_D$ and $\I_d$ are identity matrices of size $D$ and $d$, respectively. Since the exact inference of the generative process is in general intractable, a variational approximation of the evidence (marginal likelihood) can be used,
    \begin{align}
    \begin{split}
        \mathcal{L}_{ELBO}
          &= \mathbb{E}_{q_{\bm{\xi }}(\z|\x)}\left[\log(\Prob_{\bm{\phi}}(\x|\z))\right] \\
          &- \mathrm{KL}\left(q_{\bm{\xi }}(\z|\x)||\Prob(\z)\right) ,
        \label{eq:elbo}
    \end{split}
    \end{align}
    where $q_{\bm{\xi}}(\z|\x) = \mathcal{N}(\x | \mu_{\bm{\xi}}(\x), \I_d\sigma^2_{\bm{\xi}}(\x))$ approximates the posterior distribution $p(\z | \x)$ by two deep neural networks $\mu_{\bm{\xi}}(\x) : \ambient \rightarrow \latent$ and $\sigma_{\bm{\xi}}(\x)): \ambient \rightarrow \R^d_+$.
    The posterior distribution $p_{\bm{\xi }}(\z | \x)$ is called \emph{inference} or \emph{encoder} distribution, while the generative distribution $p_{\bm{\phi}}(\x|\z)$ is known as the \emph{generator} or \emph{decoder}. 
    In the next subsection, we use a VAE to learn a skill-specific Riemannian manifold from human demonstrations.  
    \begin{SCfigure*}
    	\includegraphics[width=0.7\textwidth]{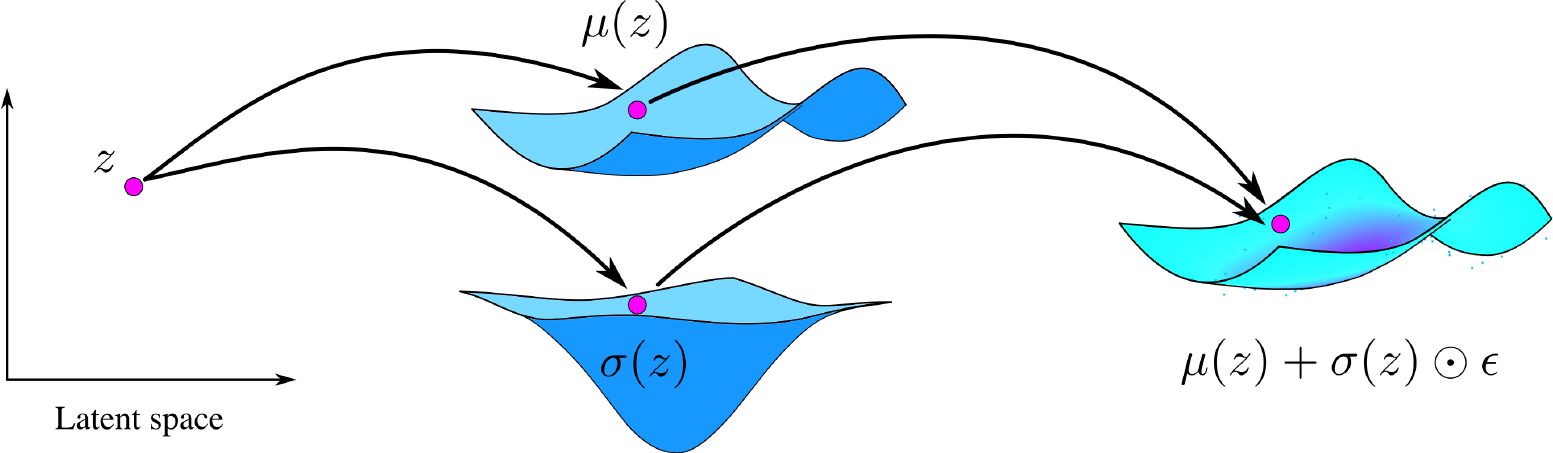}
    	\caption{In a Gaussian VAE, samples are generated by a random projection of the manifold jointly spanned by $\mu$ and $\sigma$.}
    	\label{fig:Noisy_manifold}
    \end{SCfigure*}

\subsection{Riemannian Manifolds}
    In differential geometry, Riemannian manifolds are referred to as curved $d$-dimensional continuous and differentiable surfaces characterized by a Riemannian metric~\citep{Lee18Riemann}.
    This metric is characterized by a family of smoothly varying positive-definite inner products acting on the tangent spaces of the manifold, which locally resembles the Euclidean space $\R^d$. 
    In this paper, we use the mapping function $f$ to represent a manifold $\Manifold$ immersed in the ambient space $\ambient$ defined as,
    \begin{align}
      \Manifold = f(\latent) \quad \mathrm{with} \quad f: \latent \rightarrow \ambient,
      \label{eq:mapping}
    \end{align}
    where $\latent$ and $\ambient$ are open subsets of Euclidean spaces with $\dim{\latent} < \dim{\ambient}$.
    
    An important operation on Riemannian manifolds is the computation of the length of a smooth curve $\curve: [0, 1] \rightarrow \latent$, defined as,
    \begin{align}
      \Length_{\curve}  &= \int_0^1 \| \partial_t f(\curve(t)) \| \mathrm{d}t .
      \label{eq:length}
    \end{align}
    This length can be reformulated using the chain rule as,
    \begin{align}
      \Length_{\curve}  &= \int_0^1 \sqrt{\dot{\curve}(t)^{\trsp} \Metric(c(t)) \dot{\curve}(t)} \mathrm{d}t,
      \label{eq:length_chain_rule}
    \end{align}
    where $\Metric$ and $\dot{\curve}_t = \partial_t \curve_t$ are the Riemannian metric and curve derivative, respectively. 
    Note that the Riemannian metric corresponds to, 
    \begin{equation}
        \Metric(\z) = \Jac_f(\z)^{\trsp} \Jac_f(\z) .
        \label{eq:RiemMetric}
    \end{equation}
    Here, $\Jac_f(\z)$ is the Jacobian of the mapping function $f$. 
    This metric can be used to measure local distances in $\latent$. 
    The shortest path on the manifold, also known as the geodesic, can be computed given the curve length in Eq.~\eqref{eq:length_chain_rule}. 
    Geodesics on Riemannian manifolds can be seen as a generalization of straight lines in Euclidean space. 
    However, geodesics might not be unique, e.g. great circles on the sphere manifold. 
    Later, we demonstrate that calculating geodesics on a learned Riemannian manifold can be leveraged to recover demonstrated motion patterns. It should be noted that geodesics have recently been utilized as solutions of trajectory optimizers for quadrotor control~\cite{ScannellTrajectory2021}.

\subsection{Learning Riemannian Manifolds with VAEs}
    \label{sec:vae_manifold}
    In this subsection, we examine the link between VAEs and Riemannian geometry. 
    To begin, we first define the VAE generative process of Eq.~\eqref{eq:vae_gen} as a stochastic function,
    \begin{equation}
        f_{\bm{\phi}}(\z) = \mu_{\bm{\phi}}(\z) + \operatorname{diag}(\epsilon)\sigma_{\bm{\phi}}(\z), \quad \epsilon \sim \mathcal{N}(\bm{0}, \I_D) .
        \label{eq:StochasticF}
    \end{equation}
    where $\mu_{\bm{\phi}}(\z)$ and $\sigma_{\bm{\phi}}(\z)$ are decoder mean and variance neural networks, respectively. 
    Also, $\operatorname{diag}(\cdot)$ is a diagonal matrix, and $\I_D$ is a $D \times D$ identity matrix. 
    The above formulation is referred to as the reparameterization trick~\citep{kingma:autoencoding}, which can be interpreted as samples generated out of a random projection of a manifold jointly spanned by $\mu_{\bm{\phi}}$ and $\sigma_{\bm{\phi}}$, as depicted in Fig.~\ref{fig:Noisy_manifold}.
    
    Riemannian manifolds may arise from mapping functions between two spaces as in Eq.~\eqref{eq:mapping}. 
    As a result, Eq.~\eqref{eq:StochasticF} may be seen as a stochastic version of the mapping function of Eq.~\eqref{eq:mapping}, which in turn defines a Riemannian manifold~\citep{Hauberg:OnlyBS}. 
    We can now write the stochastic form of the Riemannian metric of Eq.~\eqref{eq:RiemMetric}. 
    To do so, we first recast the stochastic function Eq.~\eqref{eq:StochasticF} as follows~\citep{eklund:arxiv:2019},
    \begin{align}
      f_{\bm{\phi}}(\z) &= \begin{pmatrix} \I_D, & \operatorname{diag}(\epsilon) \end{pmatrix} \begin{pmatrix} \mu_{\bm{\phi}}(\z) \\ \sigma_{\bm{\phi}}(\z) \end{pmatrix}
         = \bm{P}\;g(\z) ,
    \end{align} 
    where $\bm{P}$ is a random matrix, and $g(\z)$ is the concatenation of $\mu_{\bm{\phi}}(\z)$ and $\sigma_{\bm{\phi}}(\z)$. Therefore, the VAE can be seen as a random projection of a deterministic manifold spanned by $g$. 
    Given that this stochastic mapping function is defined by a combination of mean $\mu_{\bm{\phi}}(\z)$ and variance $\sigma_{\bm{\phi}}(\z)$, the metric is likewise based on a mixture of both as follows,
    \begin{equation}
        \bar{\Metric}(\z) = \Jac_{\mu_{\bm{\phi}}}(\z)^{\trsp} \Jac_{\mu_{\bm{\phi}}}(\z) + \Jac_{\sigma_{\bm{\phi}}}(\z)^{\trsp} \Jac_{\sigma_{\bm{\phi}}}(\z) .
        \label{eq:VAE_metric}
    \end{equation}
    where $\Jac_{\mu_{\bm{\phi}}}(\z)$, $\Jac_{\sigma_{\bm{\phi}}}(\z)$,  $\Jac_{\mu_{\bm{\phi}}}(\z)^{\trsp}$, and $\Jac_{\sigma_{\bm{\phi}}}(\z)^{\trsp}$ are respectively the Jacobian of $\mu_{\bm{\phi}}(\z)$ and $\sigma_{\bm{\phi}}(\z)$ and their corresponding transpose evaluated at $\z \in \latent$, with $\latent$ being the VAE low-dimensional latent space.

    Notably, the decoder variance network $\sigma_{\bm{\phi}}(\z)$ approximates the data uncertainty, which plays a critical role in the metric Eq.~\eqref{eq:VAE_metric} by associating low values to regions with a high number of data points and vice-versa. 
    Indeed, omitting this element results in a low-curvature manifold geometry~\citep{Hauberg:OnlyBS}.
    For example, \citet{Shao:TheRiemannianGeometry} suggests a similar technique that learns Riemannian manifolds using generative models that do not model data uncertainty, resulting in low-curvature manifolds and often straight lines as geodesics.
    As explained earlier, the Riemannian metric is required to compute the geodesics, which conform to the geometry of the training data~\citep{Arvanitidis:LatentSO}. 
    
    In summary, we exploit the link between VAEs and Riemannian metrics for robot motion generation. 
    Specifically, we learn a Riemannian metric that describes the motion patterns observed during the demonstrations.
    These demonstrations may take place in two different ambient spaces: Task and joint space, which define the VAE architecture needed to learn the manifold of interest.
    The geodesic curves generated on the learned manifold produce robot movements that mimic the given demonstrations in the ambient space. 

\subsection{Ambient space metric}
    According to the preceding section, we leverage a VAE to learn a Riemannian metric. 
    However, there are situations where this metric needs to be changed. 
    For example, if our metric encapsulates the main patterns of the robot motion demonstrations, we may be interested in endowing the robot with obstacle-avoidance capabilities. 
    This means that we now require to reshape the manifold to take these obstacles into account. 
    To do so, we need to reshape the previously-learned metric so that the new geodesics lead to obstacle-free robot motions. 
    
    A na\"ive approach would entail retraining the VAE model with new data, which is time-consuming and data-intensive, and can only be executed offline.
    We propose to reshape the learned metric by considering problem-specific ambient metrics. 
    Note that although the definition of curve length relies on the Euclidean metric of $\ambient$, this is not a strict requirement.
    Indeed, \cite{arvanitidis:arxiv:2020} argued that there is value in giving the ambient space a manually-defined \emph{Riemannian} metric and including that into the definition of curve length. 
    The resulting metric can then be used to compute the curve length as,
    \begin{align}
        \Length_{\curve}  &= \int_0^1 \sqrt{\dot{\curve}(t)^{\trsp} \Jac_{f_{\bm{\phi}}}(c(t))^{\trsp} \Metric_{\ambient}(f_{\bm{\phi}}(c(t))) \Jac_{f_{\bm{\phi}}}(c(t)) \dot{\curve}(t)} \mathrm{d}t,
    \end{align}
    where $\Metric_\ambient$ is the ambient metric, which can now vary smoothly across $\ambient$.
    The reshaped Riemannian metric in $\latent$ is then computed as follows,
    \begin{align}
        \bar{\Metric}(\z) &= \Jac_{\mu_{\bm{\phi}}}(\z)^{\trsp} \Metric_\ambient(\mu_{\bm{\phi}}(\z)) \Jac_{\mu_{\bm{\phi}}}(\z) \nonumber\\
        &+ \Jac_{\sigma_{\bm{\phi}}}(\z)^{\trsp} \Metric_\ambient(\mu_{\bm{\phi}}(\z)) \Jac_{\sigma_{\bm{\phi}}}(\z) .
        \label{eq:ambient_VAE_metric}
    \end{align}
    Given this metric, we can compute geodesics that are repelled from certain regions of the ambient space $\ambient$ by increasing the value of the ambient metric $\Metric_\ambient$. 
    We demonstrate how this metric reshaping method is leveraged to generate obstacle-free robot motions in Section~\ref{sec:experiments}.

%% file: Sections/Approach.tex
\section{Riemannian manifold learning}

\label{sec:riemannian_manifolds}

In this section, we describe how learning complex robot motion skills from demonstrations can be treated from a Riemannian manifold perspective. 
Unlike previous works~\cite{Havoutis:MotionPlanningManifold13,Li:TaskManifoldConstrainedManip18}, where skill manifolds are built from locally smooth manifold learning~\cite{Dollar07:LSML}, we leverage a Riemannian formulation. 
We develop a model that has enough capacity to learn and synthesize the relevant patterns of a motion while being flexible enough to adapt to new conditions (e.g. dynamic obstacles). 
In this context, related approaches such as geometric control methods build on the geometric properties of a system in the design of control laws. These methods are particularly useful in situations where the dynamics of the system are complex and difficult to model accurately~\citep{GeometricControl2004Bullo}.

On a related note, our approach shares some conceptual aspects with optimal control theory. 
Broadly speaking, optimal control aims at finding optimal control inputs that minimize a specific cost functional~\citep{OptimalControl1970Kirk}. 
The connection between inverse optimal control and Riemannian manifold learning is that the learned Riemannian metric can be understood as the cost function that the human demonstrator is optimizing. 
More specifically, the optimal expert demonstrations of a motion are used to learn a Riemannian metric that defines the optimal or shortest path (i.e., geodesics) that minimizes the functional in Eq.~\eqref{eq:length_chain_rule}. 
In comparison, our method differs conceptually from the aforementioned control techniques, as it leverages geodesic on a learned Riemannian manifold for robot motion generation. While all approaches aim to achieve precise and efficient control or planning for mechanical systems, they do so through different perspectives.
We next describe how we use VAEs to access a low-dimensional learned manifold of the demonstrations to learn an ambient space Riemannian metric. 

As mentioned previously, this metric is exploited to reconstruct robot motions in both task space $\mathbb{R}^3 \times \mathcal{S}^3$ and joint space $\R^\eta$. 
In addition, we discuss the design of the corresponding VAE for each ambient space as well as the formulation of the corresponding Riemannian metric.
In Section~\ref{sec:geodesic_motion}, we explain how we exploit these learned metrics to generate robot motion trajectories using geodesics.

\subsection{Task space $\mathbb{R}^3 \times \mathcal{S}^3$} 
    To begin, we focus on learning motion skills characterized by full-pose end-effector trajectories, where each pose is represented in $\mathbb{R}^3 \times \mathcal{S}^3$.
    Before exploiting the VAE to compute the Riemannian metric, we must ensure its capability to properly learn and reconstruct full-pose end-effector states, i.e.\ position $\x~\in~\mathbb{R}^3$ and orientation $\q~\in~\mathcal{S}^3$, while accounting for specific properties of the data, such as quaternions antipodality.
    To do so, we propose a VAE architecture that models the joint density of the robot end-effector state.
    Our model retains the usual Gaussian prior $p(\z) = \mathcal{N}(\z | \bm{0}, \I_d)$, but modifies the generative distribution $p_{\bm{\phi}, \bm{\psi}}(\x, \q | \z)$.
    Specifically, we assume that position and orientation are conditionally independent,  
    \begin{align}
        p_{\bm{\phi}, \bm{\psi}}(\x, \q | \z) = p_{\bm{\phi}}(\x | \z) p_{\bm{\psi}}(\q | \z) ,
    \end{align}
    where the latent variable $\z$ captures the correlation between position and quaternion data. 
    Next, we describe how each conditional distribution is parameterized and learned. 
    
    \subsubsection{Position encoding in $\mathbb{R}^3$:}
    To model the conditional distribution of end-effector positions $\x$, we opt for simplicity and choose this to be Gaussian,
    \begin{align}
      p_{\bm{\phi}}(\x | \z) &= \mathcal{N}(\x | \mu_{\bm{\phi}}(\z), \I_3 \sigma^2_{\bm{\phi}}(\z)) ,
    \end{align}
    where $\mu_{\bm{\phi}}$ and $\sigma_{\bm{\phi}}$ are neural networks parametrized by $\bm{\phi}$.
    %
    %
    This is a somewhat simplistic model as the Gaussian model will assign probability mass outside the workspace of the robot, i.e.\@  physically infeasible robot states. With simplicity in mind, we disregard this concern as we only optimize the associated likelihood, and therefore never sample infeasible states. However, if we were to use the VAE as a fully generative model, this likelihood should be replaced by a more elaborated model that accounts for physical properties.

\subsubsection{Quaternion encoding in $\mathcal{S}^3$:}
    On a robot motion trajectory, each position is paired with an orientation, and together they define the full pose of the end-effector. 
    There are several representations for the end-effector orientation, for example, Euler angles, rotation matrices, and unit quaternions. 
    Euler angles and rotation matrices are widely used for their simplicity and intuitiveness, however, the former suffer from gimbal lock~\citep{Hemingway2018:gimballock} which makes them an inadequate orientation parametrization, and the latter are a redundant representation requiring a high number of parameters.
    
    Unit quaternions, on the other hand, are a convenient way to represent orientations since they are compact, not redundant, and prevent gimbal locks. 
    Also, they provide strong stability guarantees in closed-loop orientation control~\citep{Camarillo08:quaternions}, and they have been recently exploited in complicated robotic tasks learning~\citep{Rozo2020:SkillsSeq}, and for data-efficient robot control tuning~\citep{Jaquier2019:GaBO} using Riemannian-manifold formulations. 
    We choose to represent orientations $\q$ as a unit quaternion, such that $\q \in \mathcal{S}^3$ with the additional antipodal identification that $\q$ and $-\q$ correspond to the same orientation. 
    Formally, a unit quaternion $\q$ lying on the surface of a $3$-sphere $\mathcal{S}^3$ can be represented using a $4$-dimensional unit vector $\q~=~[q_w, q_x, q_y, q_z]$, where the scalar $q_w$ and vector $(q_x, q_y, q_z)$ represent the real and imaginary parts of the quaternion, respectively. 
    To cope with antipodality, one could model $\q$ as a point in a projective space, but for simplicity we let $\q$ live on the unit sphere $\mathcal{S}^3$. 
    We then choose a generative distribution $p_{\bm{\psi}}(\q | \z)$ such that $p_{\bm{\psi}}(\q | \z)~=~p_{\bm{\psi}}(-\q | \z)$.
    In other words, the quaternions $\q$ and $-\q$ are considered to be antipodal: they lie on diametrically opposite points on the $3$-sphere while representing the same orientation. 
    
    To formulate a suitable distribution $p_{\bm{\psi}}(\q | \z)$ over $\mathcal{S}^3$, we leverage the von Mises-Fischer (vMF) distribution, which is merely an isotropic Gaussian constrained to lie on the unit sphere~\citep{Sra18:DirectionalStats}. 
    This distribution is described by a mean direction $\bm{\mu}$ with $\left \| \bm{\mu} \right \| = 1$, and a concentration parameter $\kappa \ge 0$.
    The vMF density function is defined as,
    \begin{align}
      \vMF(\q | \bm{\mu}, \kappa) = C_{D}(\kappa) \exp\left({\kappa\bm{\mu}^{\trsp} \q}\right) ,
      \qquad \| \bm{\mu} \| = 1,
      \label{eq:vmf_density}
    \end{align}
     where $C_{D}$ is the normalization constant
    \begin{align}
    C_{D}({\kappa}) = \frac{\kappa^{\frac{D}{2}-1}}{(2\pi)^{\frac{D}{2}} \mathit{I}_{\frac{D}{2}-1}(\kappa)} ,
    \end{align}
    with $\mathit{I}_{\frac{D}{2}-1}(\kappa)$ being the modified Bessel function of the first kind. 
    Like the Gaussian, from which the distribution was constructed, the von Mises-Fischer distribution is unimodal. 
    To build a distribution that is antipodal symmetric, i.e.\ $p_{\bm{\psi}}(\q | \z) = p_{\bm{\psi}}(-\q | \z)$, we define a mixture of antipodal vMF distributions~\citep{hauberg:tpami:grassmann},
    \begin{align}
      p_{\bm{\psi}}(\q | \z) &= \frac{1}{2} \vMF(\q | \bm{\mu}_{\bm{\psi}}(\z), \kappa_{\bm{\psi}}(\z)) \nonumber\\
      &+ \frac{1}{2} \vMF(\q | -\bm{\mu}_{\bm{\psi}}(\z), \kappa_{\bm{\psi}}(\z)) ,
    \end{align}
    where $\bm{\mu}$ and $\kappa$ are parametrized as neural networks. 
    This mixture model is conceptually similar to a Bingham distribution~\citep{Sra18:DirectionalStats}, but is easier to implement numerically.

\subsubsection{Variational inference:}
    To train the VAE, we maximize an adapted evidence lower bound (ELBO) Eq.~\eqref{eq:elbo}, defined as
    \begin{align}
       \label{eq:final_elbo}
       \Loss_{ELBO} &= \beta_1 \Loss_\x +  \beta_2 \Loss_\q - \mathrm{KL}\left(q_{\bm{\xi}}(\z|\x)||\Prob(\z)\right) ,\\
       \Loss_\x &=  \mathbb{E}_{q_{\bm{\xi}}(\z|\x)}\left[\log p_{{\bm{\phi}}}(\x|\z) \right] , \\
       \Loss_\q &=  \mathbb{E}_{q_{\bm{\xi}}(\z|\x)}\left[\log p_{{\bm{\psi}}}(\q|\z) \right] ,
    \end{align}
    where $\x \in \R^3$ and $\q \in \Sph^3$ represent the position and orientation of the end-effector, respectively.
    The scaling factors $\beta_1>0$ and $\beta_2>0$ balance the log-likelihood of position and orientation components. This approach is inspired by the $\beta$-VAE method~\citep{betaVAELB201Higgins}, which, despite not yielding a valid lower bound on the ELBO objective function, has been shown to produce high-quality results comparable to methods that provide a valid lower bound, thus it is commonly used in practice.
    Due to the quaternions antipodality, raw demonstration data may contain negative or positive values for the same orientation. 
    So, we avoid any pre-processing step of the data by considering two vMF distributions that encode the same orientation at both sides of the hypersphere. 
    Practically, we double the training data, by including $\q_n$ and $-\q_n$ for all observations $\q_n$.
    Note that as the Riemannian manifold is learned using task space data, the model is kinematics agnostic, which means that the generated motion may be used across different robots as long as the trajectory is reachable.

\subsubsection{Induced Riemannian metric:}
    Our generative process is parametrized by a set of neural networks. Specifically, $\mu_{\bm{\phi}}$ and $\sigma_{\bm{\phi}}$ are position mean and variance neural networks parameterized by $\bm{\phi}$, while $\mu_{\bm{\psi}}$ and $\kappa_{\bm{\psi}}$ are neural networks parameterized by $\bm{\psi}$ that represent the mean and concentration of the quaternion distribution. 
    Following Section~\ref{sec:vae_manifold}, the Jacobians of these functions govern the induced Riemannian metric as,
    \begin{align}
      \label{eq:pos_quat_metric}
      \Metric(\z) &= \Metric_\mu^\x(\z) + \Metric_\sigma^\x(\z) + \Metric_\mu^\q(\z) + \Metric_\kappa^\q(\z) ,
    \end{align}
    with
    \begin{align}
      \Metric_\mu^\x(\z) &= \Jac_{\mu_{\bm{\phi}}}(\z)^{\trsp} \Jac_{\mu_{\bm{\phi}}}(\z) , \\ \Metric_\sigma^\x(\z) &= \Jac_{\sigma_{\bm{\phi}}}(\z)^{\trsp} \Jac_{\sigma_{\bm{\phi}}}(\z) ,\\
      \Metric_\mu^\q(\z) &= \Jac_{\mu_{\bm{\psi}}}(\z)^{\trsp} \Jac_{\mu_{\bm{\psi}}}(\z) ,  \\ \Metric_\kappa^\q(\z) &= \Jac_{\kappa_{\bm{\psi}}}(\z)^{\trsp} \Jac_{\kappa_{\bm{\psi}}}(\z) ,
    \label{eq:OurMetric}
    \end{align}
    where $\Jac_{\mu_{\bm{\phi}}}$, $\Jac_{\sigma_{\bm{\phi}}}$, $\Jac_{\mu_{\bm{\psi}}}$, $\Jac_{\kappa_{\bm{\psi}}}$ are the Jacobians of the functions representing the position mean and variance, as well as the quaternion mean and concentration, respectively.

    In practice, we want this Riemannian metric $\Metric(\z)$ to take large values in regions with little or no data, so that geodesics avoid passing through them. 
    We achieve this by using radial basis function (RBF) networks as our variance representation, whose kernels reliably extrapolate over the whole space~\citep{Arvanitidis:LatentSO}.
    Since one of the main differences between Gaussian and von Mises-Fischer distributions is the representation of the data dispersion, the RBF network should consider a reciprocal behavior when estimating variance for positions. 
    In summary, the data uncertainty is encoded by the RBF networks representing $\sigma^{-1}_{\bm{\phi}}(\z)$ and $\kappa_{\bm{\psi}}(\z)$, which affect the Riemannian metric through their corresponding Jacobians as in Eq.~\eqref{eq:pos_quat_metric}.     

\subsection{Joint space $\mathbb{R}^{\eta}$} 
    The joint space $\mathbb{R}^{\eta}$, also known as \emph{configuration space}, is another space to represent robot motion trajectories\footnote{We did not model the robot joint space as a high-dimensional torus for simplicity. However, we showed that our approach can easily encode data on Riemannian manifolds as in the task space case.}. 
    In this space, each trajectory point is represented as the vector $\bm{\theta} = \left [\theta_1, \theta_2, \ldots, \theta_\eta \right ]~\in~\R^\eta$, where $\eta$ is the number of degrees of freedom of the robot. 
    Learning motion skills in this space is known to be challenging as it is less intuitive to provide joint-level demonstrations.
    However, being able to learn and generate joint space movements is relevant as some tasks may demand specific robot postures.
    Moreover, joint space skills can be extended to provide whole-body obstacle avoidance.
    In this context, we formulate a Riemannian robot motion learning approach to generate collision-free joint space movements. 
    
    Previously, we computed a Riemannian metric in the latent space using the VAE decoder trained on task space demonstrations.
    Similarly, a new VAE architecture can be designed to compute a Riemannian metric from joint space demonstrations. 
    We use this metric to compute geodesics and generate robot movements that resemble the demonstrations in joint space.
    This joint space approach also allows us to endow the robot with whole-body obstacle avoidance capabilities.
    By using ambient metrics, we can again reshape the learned metric to make the robot move away from obstacles in an online fashion.
    The ambient metrics exclusively use task space information of the obstacles and the robot body, in contrast to classical motion planning that often works in the configuration space.
    Note that the data manifold learned using joint space demonstrations is kinematics-dependent, meaning that the generated motion cannot be directly transferred to other robots with different kinematics.

\subsubsection{Variational inference:}
   To train the joint space VAE, we maximize a modified version of the evidence lower bound (ELBO) Eq.~\eqref{eq:elbo}, defined as
       \begin{equation} 
    \mathcal{L}_{ELBO} = \Loss_{\bm{\theta}}-\mathrm{KL}\left(q_\xi(\bm{z}_i|\bm{\theta}_i)||p(z)\right) ,
    \label{equ:final_ELBO_joint}
    \end{equation}
    \noindent
    with,
    \begin{align*}
    \Loss_{\bm{\theta}} = \mathbb{E}_{q_\xi(\bm{z}_i|{\bm{\theta}}_i)}&\left [ p_{\mathcal{X}}(f_{\textrm{FK}}(\bm{\theta})|\bm{z}_i)\right ] , \\
    = \mathbb{E}_{q_\xi(\bm{z}_i|{\bm{\theta}}_i)}&\left[ \log(p_\Theta(\bm{\theta}|\bm{z}_i)) -  \log(\mathcal{V})\right] ,
    \end{align*}
    where $p_\Theta(\bm{\theta}|\bm{z}_i)$ and $p_{\mathcal{X}}(f_{\textrm{FK}}(\bm{\theta})|\bm{z}_i)$ are the estimated conditional densities in the joint space $\Theta$ and task space $\ambient$, respectively. 
    Also, $\mathcal{V}$ is the volume measure defined as 
    \begin{align}
        \mathcal{V} =\sqrt{(\det(\Jac_{f_{\textrm{FK}}}^{\mathsf{T}}(\mu_{\alpha}(\bm{z}_i)) \Jac_{f_{\textrm{FK}}}(\mu_{\alpha}(\bm{z}_i)))} ,
        \label{eq:volume_joint}
    \end{align}
    where $\Jac_{f_{\textrm{FK}}}$ is the Jacobian of the forward kinematics $f_{\textrm{FK}}$ given the joint configuration estimated by the VAE decoder $\mu_{\alpha}$. 
    Furthermore, the generative distribution $p_\Theta(\bm{\theta}|\bm{z}_i)~=~\mathcal{N}(\mu_{\alpha}(\bm{z}_i), \I_{\eta}\sigma_{\alpha}(\bm{z}_i)^2)$ is parameterized by the VAE decoder mean $\mu_{\alpha}(\bm{z}_i)$ and variance $\sigma_{\alpha}(\bm{z}_i)$ networks. 
    Note that the new ELBO formulation in Eq.~\eqref{equ:final_ELBO_joint} leverages the change of variable theorem~\citep{MathForML2020:Deisenroth} to transform probability densities from joint to task space. 
    As a result, the VAE is still trained using task space information, while the given demonstration trajectories are defined in joint space.
    This is motivated by the fact that most robot skills may still depend on task space variables (e.g. the manipulated objects pose), despite the same skill is also required to imitate particular robot postures.
    
    As we are interested in whole-body obstacle avoidance, we can leverage the forward kinematics model to access the Cartesian position of different points on the robot (e.g., joint locations). 
    Therefore, we use a set of $M$ forward kinematic functions $f_{\textrm{FK}}^{1:M}(\bm{\mu_{\alpha}}^{1:\eta})$, where $\bm{\mu_{\alpha}}^{1:n}$ is $n=1, \hdots, \eta$ elements of the joint-values vector $\mu_{\alpha}(\bm{z})$, and $M$ is the number of considered points on the robot.
    Note that for certain points on the robot structure, the forward kinematics only needs a subset of the joint values.
    For simplicity, we consider $M$ to be equal to the number of robot joints plus the end-effector (i.e. $M=\eta+1$). 
    Then, the full forward kinematic function $f_{\textrm{FK}}$ is defined as, 
    \begin{align*}
    f_{\textrm{FK}}(\bm{\mu_{\alpha}}) =& 
    \begin{bmatrix} 
    f_{\textrm{FK}}^1(\bm{\mu_{\alpha}}^{1}),
    \hdots,
    f_{\textrm{FK}}^M(\bm{\mu_{\alpha}}^{1:\eta}),
    f_{\textrm{FK}}^{ee}(\bm{\mu_{\alpha}}^{1:\eta})
    \end{bmatrix}^\trsp,\\
    =& \begin{bmatrix} 
    \bm{p}_{1},
    \hdots,
    \bm{p}_{M},
    \bm{p}_{\textrm{ee}}, 
    \bm{q}_{\textrm{ee}}
    \end{bmatrix}^\trsp,
    \end{align*}
    where given the joint value vector $\bm{\mu_{\alpha}}^{1:n}$ as input, all the functions compute the corresponding position $\bm{p}_{m}$ of the $m$-th point on the robot, except the last function $f_{\textrm{FK}}^{ee}$ which also provides both the position $\bm{p}_{\textrm{ee}}$ and the orientation $\bm{q}_{\textrm{ee}}$ of the end-effector.

    Furthermore, the volume measure $\mathcal{V}$ in Eq.~\eqref{eq:volume_joint} uses the Jacobian of the full forward kinematics function, which is defined as,
    \begin{align*}
    \Jac_{f_{\textrm{FK}}(\bm{\mu_{\alpha}})} =& 
    \begin{bmatrix} 
    \Jac_{\bm{p_{1}}},
    \hdots,
    \Jac_{\bm{p}_M},
    \Jac_{\bm{p}_{\textrm{ee}}},
    \Jac_{\bm{q}_{\textrm{ee}}}
    \end{bmatrix}^\trsp,
    \end{align*}
    where $\Jac_{\bm{p}_i}$ and $\Jac_{\bm{q}_i}$ are the linear and angular components of the corresponding Jacobians.

\subsubsection{Induced Riemannian metric:}
    With the new integrated forward kinematic layer, we can calculate a pullback metric that directly uses task space information. 
    This adds an additional step in the formulation of the Riemannian metric, which now requires the Jacobian of the forward kinematics $\Jac_{f_{\textrm{FK}}}$ as well as the Jacobians of the VAE decoder $\Jac_{\mu_{\bm{\alpha}}}$ and $\Jac_{\sigma_{\bm{\alpha}}}$, computed from the mean and variance decoder networks. 
    Using these two Jacobians the metric can be defined as,
    \begin{equation}
    \Metric^{\bm{\theta}}(\z) = \Metric^{\bm{\theta}}_{\mu_{\bm{\alpha}}}(\z) + \Metric^{\bm{\theta}}_{\sigma_{\bm{\alpha}}}(\z)
    \label{eq:Joint_space_VAE_Metric}
    \end{equation}
    with,
    \begin{align*}
        \Metric^{\bm{\theta}}_{\mu_{\bm{\alpha}}}(\z) &= \left (\Jac_{f_{\textrm{FK}}}(\mu_{\bm{\alpha}}(\z)) \Jac_{\mu_{\bm{\alpha}}}(\z) \right ) ^ \trsp \left (\Jac_{f_{\textrm{FK}}}(\mu_{\bm{\alpha}}(\z)) \Jac_{\mu_{\bm{\alpha}}}(\z) \right ) ,\\
        \Metric^{\bm{\theta}}_{\sigma_{\bm{\alpha}}}(\z) &= \left (\Jac_{f_{\textrm{FK}}}(\mu_{\bm{\alpha}}(\z)) \Jac_{\sigma_{\bm{\alpha}}}(\z) \right ) ^ \trsp \left (\Jac_{f_{\textrm{FK}}}(\mu_{\bm{\alpha}}(\z)) \Jac_{\sigma_{\bm{\alpha}}}(\z) \right ) .
    \end{align*}
    Similarly to our Riemannian metric in task space, this new metric $\Metric^{\bm{\theta}}(\z)$ takes large values in regions with little or no data so that geodesics avoid passing through them. 
    Therefore, geodesic curves generated via Eq.~\eqref{eq:Joint_space_VAE_Metric} allow us to reproduce joint space robot skills. 
    
\section{Geodesic motion skills}
\label{sec:geodesic_motion}
As explained previously, geodesics follow the trend of the data, and they are here exploited to reconstruct motion skills that resemble human demonstrations. 
In this section, we describe geodesic computation in both settings, namely, where the VAE is trained on task space or joint space trajectories. 
Moreover, we explain how new geodesic paths, that avoid obstacles on the fly, can be obtained by metric reshaping.
In particular, we exploit ambient space metrics defined as a function of the obstacle's configuration to locally deform the original learned Riemannian metric. kinematic

Last but not least, our approach can encode multiple-solution skills, from which new hybrid trajectories (not previously shown to the robot) can be synthesized. To elaborate, a multiple-solution task requires the robot to combine multiple demonstrated solutions, i.e. trajectories, in order to complete it. By being able to combine multiple solutions to generate a mixed novel trajectory, the robot can find a novel solution to a task despite that not a single demonstration provides a complete solution on its own. This can be particularly useful in complex or dynamic environments where the robot needs to be able to adapt and respond to changing conditions. 
We elaborate on each of these features in the sequel. 

    \begin{figure}[t]
    \centering
    \begin{subfigure}{\linewidth}
       \includegraphics[width=1\linewidth]{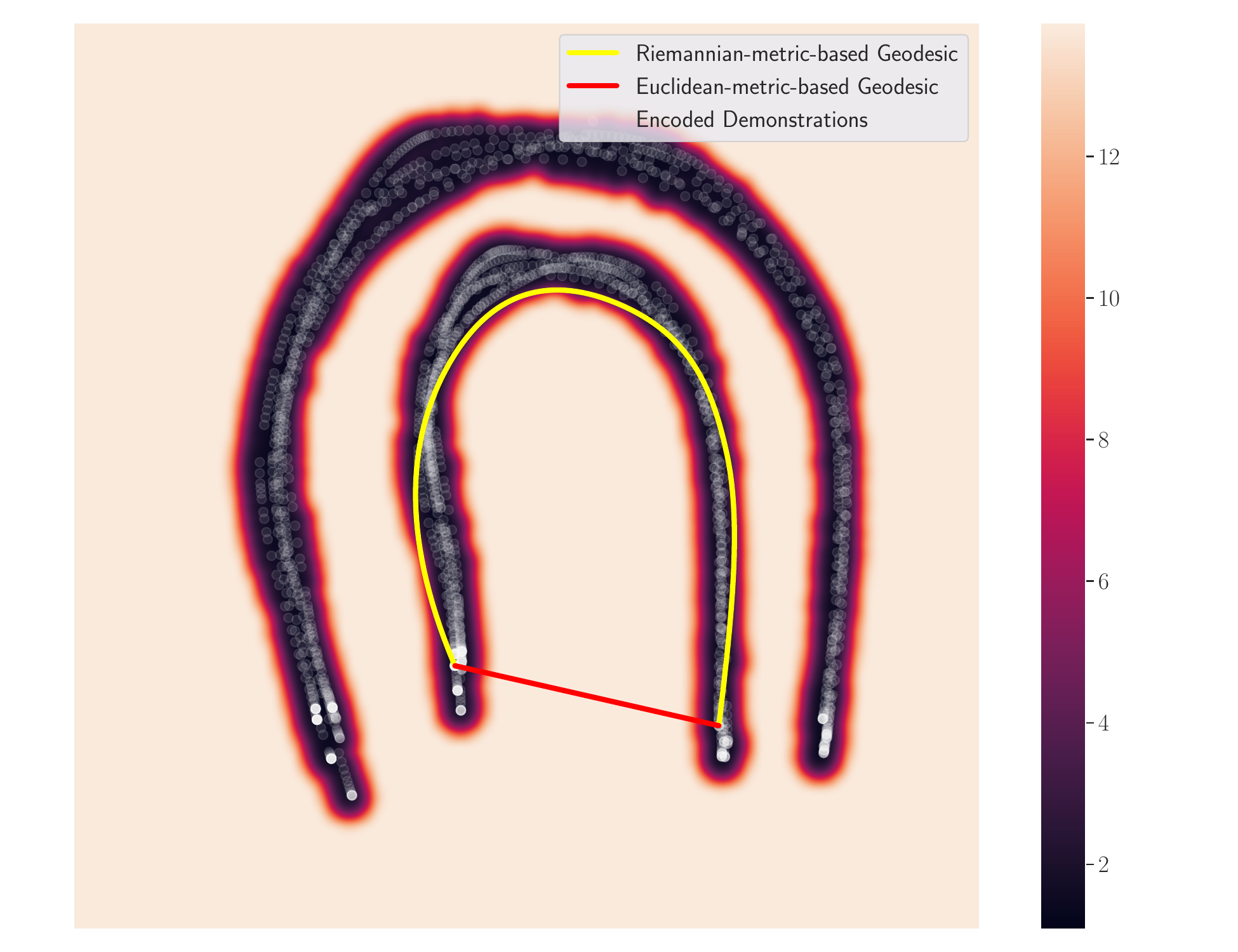}
    \end{subfigure}
    \begin{subfigure}{\linewidth}
       \includegraphics[width=1\linewidth]{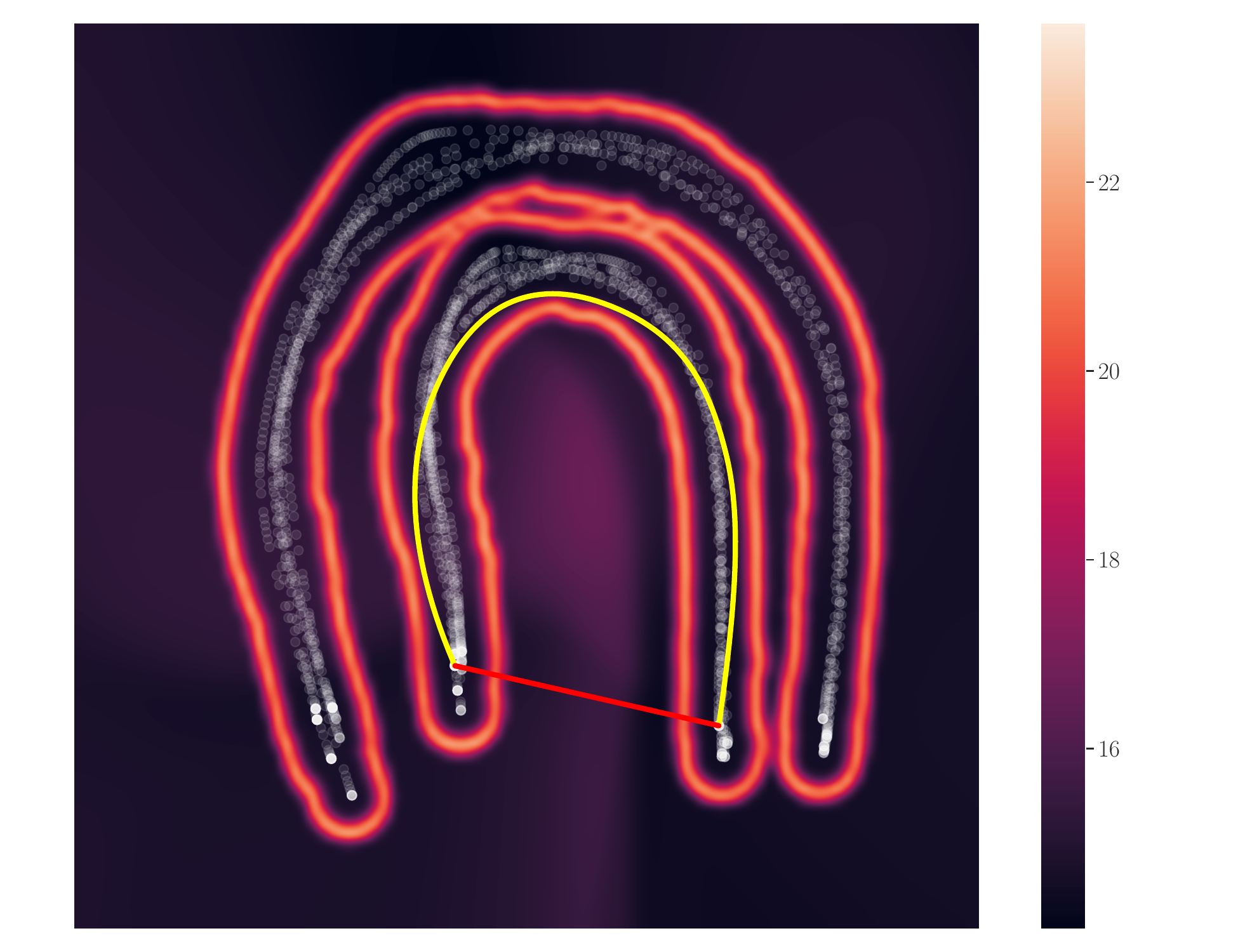}
    \end{subfigure}
    \caption{ \emph{Top}: The variance measure, \emph{bottom}: The magnification factor of the Riemannian manifold learned from trajectories based on $\mathsf{J}$ and $\mathsf{C}$ English alphabet characters defined in $\R^2 \times \mathcal{S}^2$. The semi-transparent white points depict the encoded training set, and the yellow curve depicts the geodesic in the latent space. The resulting manifold is composed of two similar clusters due to the antipodal encoding of the quaternions, where each cluster represents one side of the hyper-sphere. The yellow and red curves show the geodesics computed based on Riemannian and Euclidean metrics, respectively.}
    \label{fig:Toy_example}
    \end{figure}
    
\subsection{Generating motion:}
    Robot motion generation techniques that leverage demonstrations aim to replicate the demonstrated movement patterns with a high degree of accuracy, reliability, and efficiency. This is achieved by exploiting the demonstrations as a source of information about the range of (manifold of) possible movements that the robot should perform. The robot uses this information to generate similar movements on its own.

    One way to represent the observed movement patterns mathematically is through the use of Riemannian manifolds. Riemannian manifolds allow us to describe data sub-spaces (defined by the demonstrations) within a larger space and provide analytical tools for calculating distances and angles within these sub-spaces. In the context of robot motion generation, a Riemannian manifold can be used to characterize the region of the state space covered by the demonstrations.
    
    In this paper, the Riemannian metrics that describe the structure of these manifolds are highly influenced by the data uncertainty and they are used to find the path between two points on the manifold, the so-called geodesic. In the context of Riemannian manifolds, geodesics are the curves that minimize the distance between two points on the manifold. It is important to keep in mind that the distance measurement is done on the manifold, using the distance metric of the physical space where the robot moves. This means that geodesics on a Riemannian manifold represent the most efficient paths through the space represented by the manifold. 
    Given that Riemannian manifolds represent the sub-space within which the demonstrations reside, it follows that the geodesics are equivalent to the optimal motion trajectories.
   
    Specifically, the Riemannian metrics Eq.~\eqref{eq:VAE_metric} and Eq.~\eqref{eq:Joint_space_VAE_Metric} tell us that geodesics are penalized for crossing through regions where the VAE predictive uncertainty grows.
    This implies that if a set of demonstrations follows a circular motion pattern, geodesics starting from arbitrary points on the learned manifold will also generate a circular motion (see Fig. \ref{fig:Teaser}). 
    This behavior is due to the way that the metric $\Metric$ is defined: Our Riemannian metric $\Metric$ is characterized by low values where data uncertainty is low (and vice-versa). 
    Since the geodesics minimize the energy of the curve between two points on $\Manifold$, which is a function of $\Metric$, they tend to stay on the learned manifold and avoid outside regions.
    
    This property follows the common characteristics of motion generation techniques to reproduce motion learned from demonstrations and, makes us suggest that geodesics form a natural motion generation mechanism. 
    Note that when using a Euclidean metric (i.e., an identity matrix), geodesics correspond to straight lines. 
    Such geodesics certainly neglect the data manifold geometry.
    
    Noted that geodesics do not typically follow a closed-form equation on these learned manifolds, and numerical approximations are required. 
    This can be done by direct minimization of curve length~\citep{Shao:TheRiemannianGeometry, kalatzis:icml:2020}, $\textrm{A}^*$ search~\citep{Chen2019:FastApproximateGeodesics}, integration of the associated ODE~\citep{arvanitidis:aistats:2019}, or various heuristics~\citep{chen:MetricsforDeep}.
    In this paper, we compute geodesics on $\Manifold$ by approximating them by cubic splines $\curve \approx {\omega_\lambda}(\z_{c})$,
    with $\z_c~=~\{\z_{c_0}, \ldots, \z_{c_K} \}$, where $\z_{c_k} \in \latent$ is a vector defining a control point of the spline over the latent space $\latent$. 
    Given $K$ control points, $K-1$ cubic polynomials $\omega_{\lambda_i}$ with coefficients $\lambda_{i,0}$, $\lambda_{i,1}$, $\lambda_{i,2}$, $\lambda_{i,3}$ have to be estimated to minimize its Riemannian length,
    \begin{equation}
    \label{eq:cost_geodesic}
    \Loss_{{\omega_\lambda}(\z_{c})} = \int_0^1 \sqrt{\left \langle \dot{\omega}_\lambda(\z_{c}), \Metric(\omega_\lambda(\z_{c}))\dot{\omega}_\lambda(\z_{c}) \right \rangle} \mathrm{d}t .
    \end{equation}
    The resulting geodesic $\curve$ computed in $\latent$ is used to generate the robot motion by decoding it through the VAE networks $\mu_{\bm{\psi}}$ and $\mu_{\bm{\psi}}$ or $\mu_{\bm{\alpha}}$ depending on the ambient space setting. 
    The obtained trajectory is then executed on the robot arm to reproduce the required skill. 
    In the task space setting, the decoded geodesics can be deployed on the robot using a Cartesian impedance controller or inverse kinematics. 
    In the joint space setting, the decoded geodesics can be employed directly on the robot as a joint trajectory reference to be tracked by joint position or impedance controllers.

\subsection{Geodesics in task space $\R^3 \times \Sph^3$:}
\label{sec:Geodesics_in_task_space}
\begin{figure}
      \centering
      \includegraphics[width=1.0\linewidth]{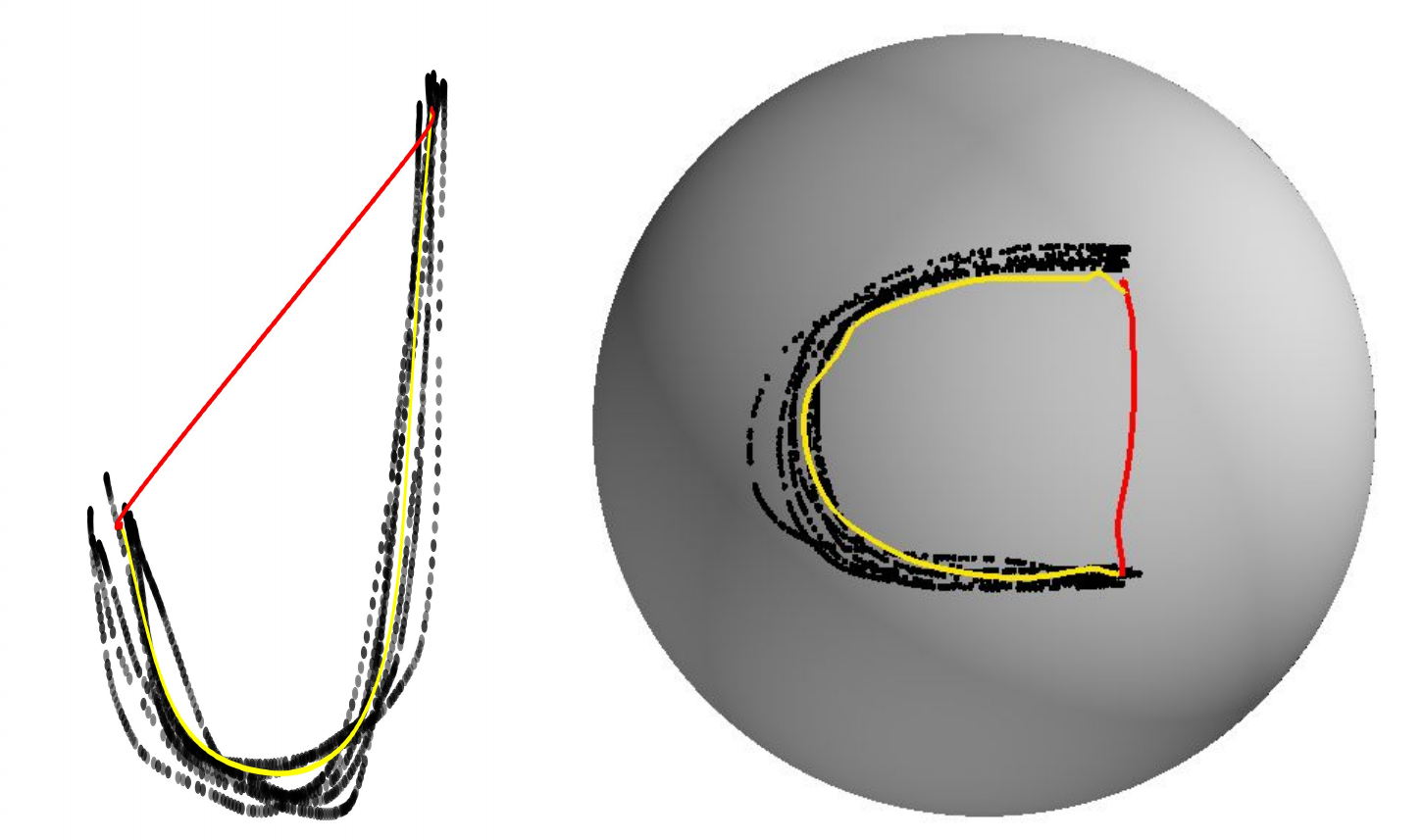}
    \caption{As an illustration, we consider synthetic data that belong to $\R^2 \times \Sph^2$. The left panel depicts the $\mathsf{J}$-shaped position data in $\R^2$ and the right panel shows the $\mathsf{C}$-shaped orientation data on $\mathcal{S}^2$. The yellow and red curves show the decoded geodesics depicted in Fig.~\ref{fig:Toy_example}, computed according to the Riemannian and Euclidean metrics, respectively.}
    \label{fig:Toy_example_data}
    \end{figure}

    In this section, we investigate the geodesic motion generation in task space. 
    To illustrate the motion generation mechanism, we consider a simple experiment where the demonstration data at each time point is confined to $\R^2 \times \Sph^2$, i.e. only two-dimensional positions and orientations are considered. 
    We artificially create position data that follows a $\mathsf{J}$ shape and orientation data that follows a $\mathsf{C}$ shape projected on the sphere (see Fig.~\ref{fig:Toy_example_data}). 
    We fit our VAE model to this dataset, and visualize the corresponding latent space in Fig.~\ref{fig:Toy_example}, where the top panel shows the latent mean embeddings of the training data with a background color corresponding to the predictive uncertainty. 
    We see low uncertainty near the data, and high otherwise. 
    The bottom panel of Fig.~\ref{fig:Toy_example} shows the same embedding but with a background color proportional to $\log\sqrt{\det\Metric}$. 
    This quantity, known as the magnification factor~\citep{Bishop1997:magfactor}, generally takes large values in regions where distances are large, implying that geodesics will try to avoid such regions. 
    In Fig.~\ref{fig:Toy_example}, we notice that the magnification factor is generally low, except on the `boundary' of the data manifold, i.e. in regions where the predictive variance grows. 
    Consequently, we observe that Riemannian geodesics (yellow curves in the figure) stay within the `boundary' and hence resemble the training data patterns. 
    In contrast, Euclidean geodesics (red curves in the figure) fail to stay in the data manifold. 
    Our proposal is to use these geodesics on the learned manifolds as our robot motion generation mechanism.
    
    Note that both panels in Fig.~\ref{fig:Toy_example} depict two distinct horseshoe-like clusters, which is a result of the antipodality of the data in $\Sph^2$. 
    More precisely, the bimodal distribution of the antipodal quaternion data is encapsulated by these two clusters in the latent space.
    In practical settings, as long as the geodesic curve does not cross across clusters (both start and goal points belong to the same cluster), the quaternion sign is unchanged. 
    We experimentally examine this in Section~\ref{sec:experiments}. 

    \begin{figure*}
      \centering
      \begin{subfigure}{.33\linewidth}
        \centering
        \includegraphics[width = \linewidth]{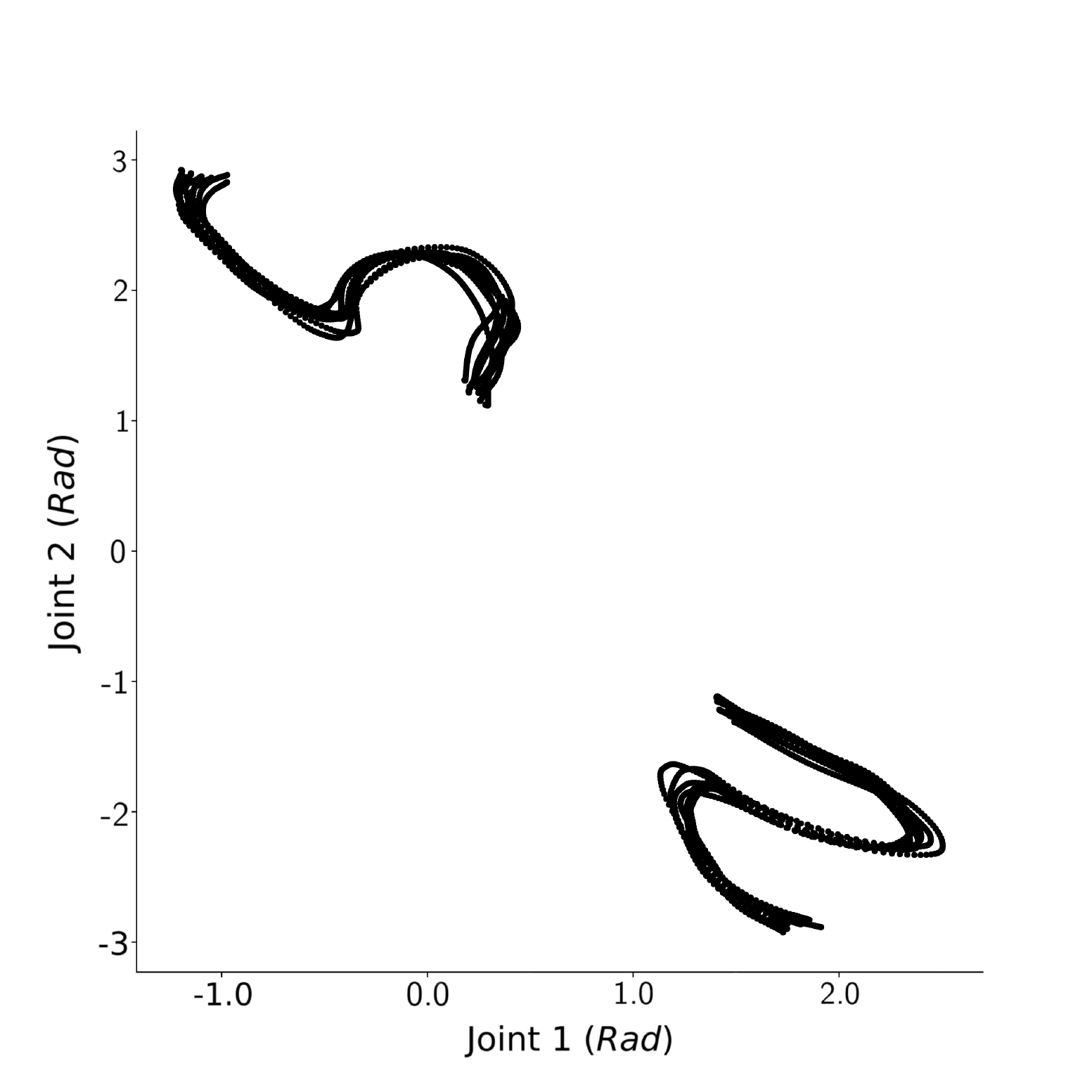}
      \end{subfigure}%
      \begin{subfigure}{.33\linewidth}
        \centering
        \includegraphics[width = \linewidth]{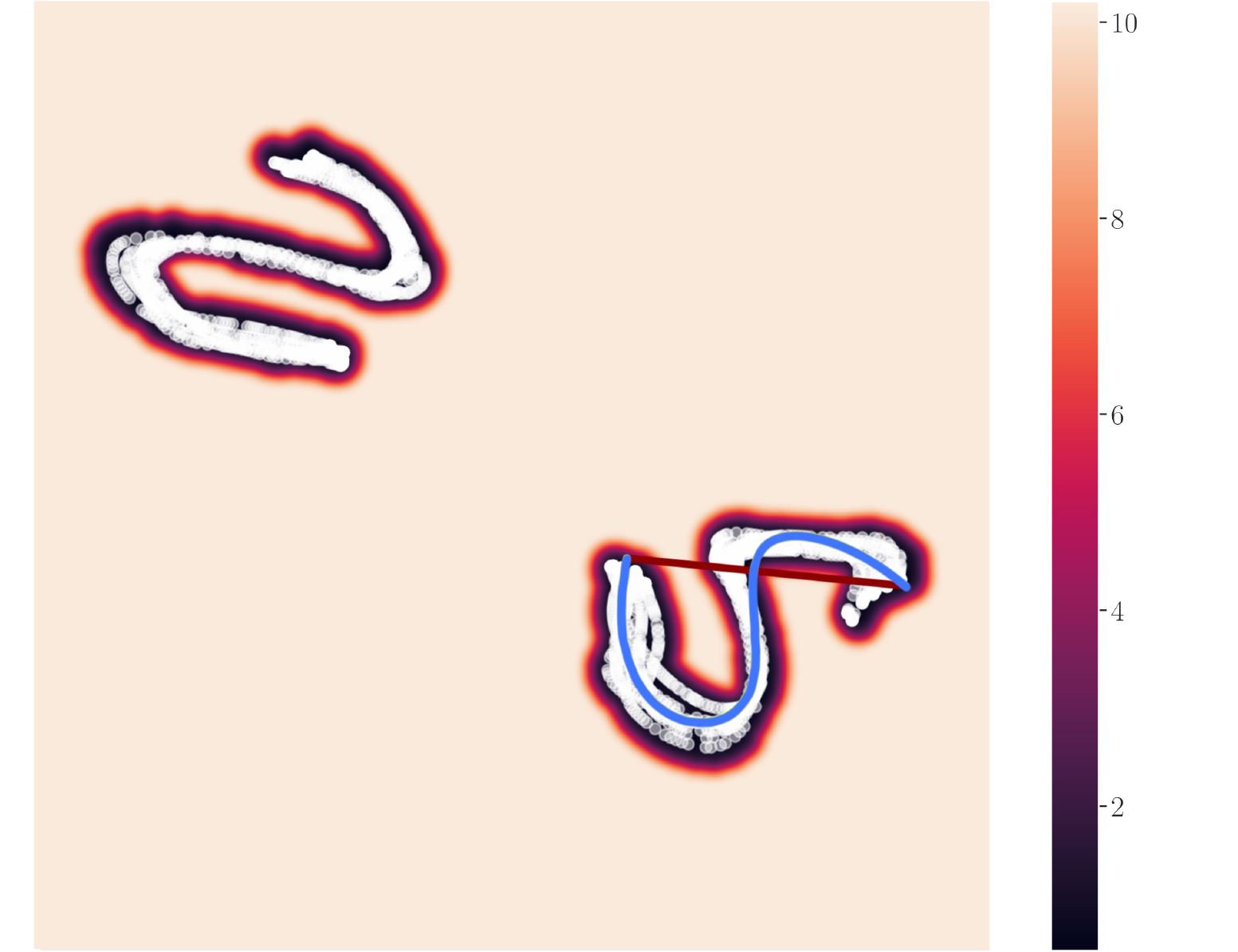}
      \end{subfigure}
        \begin{subfigure}{.33\linewidth}
        \centering
        \includegraphics[width = \linewidth]{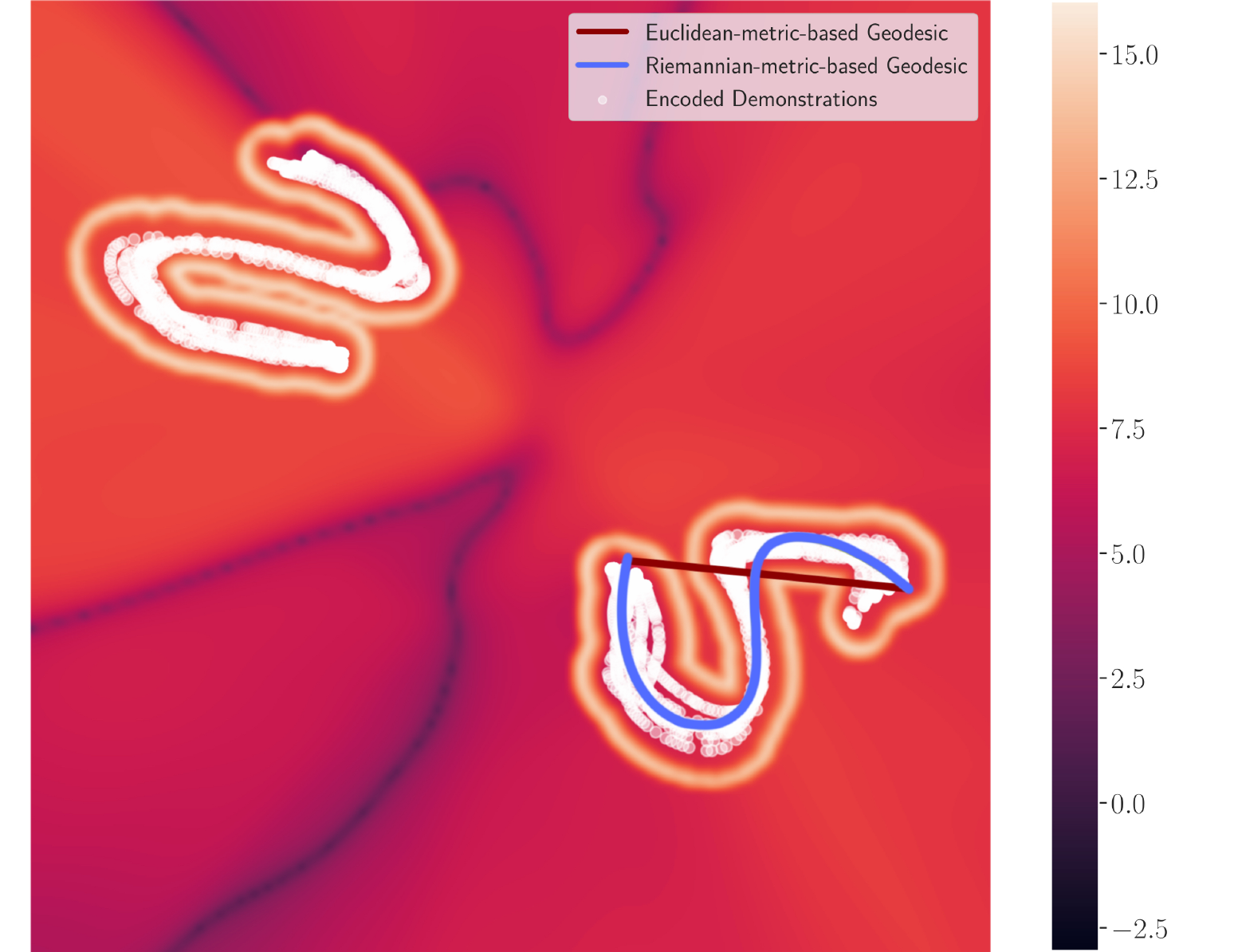}
      \end{subfigure}
      \caption{Illustration of joint space motion generation via geodesics. \emph{Left}: $\mathsf{S}$-shaped joint space demonstrations. \emph{Middle}: the resulting variance measure. \emph{Right}: The magnification factor of the learned Riemannian manifold. The semi-transparent white points depict the encoded training set and the blue curve represents the geodesic in the latent space. The resulting manifold is composed of two similar clusters due to the two different inverse-kinematics solutions for the task. The blue and red curves show the geodesics computed based on Riemannian and Euclidean metrics, respectively.}
      \label{fig:Toy_Example_Joint_Metric}
    \end{figure*}
    
\subsection{Geodesics in joint space $\R^\eta$:}
    In this section, we investigate the geodesic computation for joint space movements. 
    We use a toy example where a $2$-DOF robot arm follows an $\mathsf{S}$-shaped trajectory in task space using two different joint configurations (i.e., two different inverse-kinematics solutions), as shown in Fig.~\ref{fig:Toy_Example_Joint_Metric}--left.
    We observe two sets of demonstrations in joint space that reproduce the same end-effector movements when applied to the robot.
    We can also see in the middle and right panels of Fig.~\ref{fig:Toy_Example_Joint_Metric} the geodesics computed using the Riemannian and Euclidean metrics, depicted as blue and red curves, respectively. 
    The background of the middle panel illustrates the predictive uncertainty over the latent space $\latent$, where we again see low uncertainty near the data, and high otherwise.
    For completeness, the background in the right panel illustrates the magnification factor. 
    The latent mean embedding of the training data is depicted as semi-transparent white points. 
    Similar to the previous section, the geodesics generated using the Riemannian metric stay within the `boundary' near the training data. 
    
    Furthermore, it is easy to note that the learned manifold comprises two clusters, but unlike the previous task space example, these clusters arise from the two different joint space solutions provided in the training data.
    This indicates that the clusters in the learned manifold encapsulate the provided solutions in the demonstrations. 
    When the number of clusters grows, the geodesic has a higher chance to travel among them to find a path with minimal energy as the high-energy regions may become narrow. 
    However, unnecessary frequent switching among these clusters may often lead to jerky geodesics, therefore negatively impacting the geodesics quality, particularly in robots with a high degree of freedom (e.g. $\textrm{DOF} \ge 7$).
    Later in Section~\ref{sec:experiments}, we experimentally show that this issue can be alleviated by increasing the latent space dimensionality.

\subsection{Obstacle avoidance using ambient space metrics}
    Often human demonstrations do not include any notion of obstacles in the environment. 
    Therefore, obstacle avoidance is usually treated as a separate problem when generating robot motions in unstructured environments. 
    A possible solution to integrate both problems is to provide obstacle--aware demonstrations, where the robot is explicitly taught how to avoid known obstacles. 
    The main drawback is that the robot is still unable to avoid unseen obstacles on the fly. 

    Our Riemannian approach provides a natural solution to this problem. 
    The learned metrics in latent space Eq.~\eqref{eq:VAE_metric} and Eq.~\eqref{eq:Joint_space_VAE_Metric} measure the length of a geodesic curve under the Euclidean space of the ambient space $\ambient$. 
    We can easily modify this to account for unseen and dynamic obstacles. 
    Intuitively, we can increase the length of curves that intersect obstacles, such that geodesics are less likely to go near them. Next, we explain how obstacle avoidance can be achieved for both task space and joint space settings. 

\subsubsection{Obstacle avoidance in task space $\R^3 \times \Sph^3$:}
    Here we explain how we can reshape the learned metric to avoid obstacles in the task space setting, where only the robot end-effector is considered.
    Formally, we propose to define the ambient metric of the end-effector position to be
    \begin{align}
      \Metric_\ambient^\x(\x) &= \left( 1 + \zeta  \exp\left( \frac{-\| \bm{x} - \bm{o} \|^2}{2r^2} \right)\right)\I_3,
      \quad \bm{x} \in \mathbb{R}^3,
    \label{eq:ambient_metric_R3S3}
    \end{align}
    where $\zeta > 0$ scales the cost, $\bm{o} \in \R^3$ and $r>0$ represent the position and radius of the obstacle, respectively. 
    For the orientation component, we assume a flat ambient metric $\Metric_\ambient^\q(\x) = \I_4$. 
    Under this new ambient metric, geodesics will generally avoid the obstacle, though we emphasize this is only a \emph{soft} constraint. 
    This approach is similar in spirit to CHOMP~\citep{Ratliff2009:chomp} except our formulation works along a low-dimensional learned manifold, whose solution is then decoded to the task space of the robot.

    Under this ambient metric, the associated (reshaped) Riemannian metric of the latent space $\latent$ becomes,
    \begin{equation}
    \label{eq:obstacle_metric}
    \Metric(\z) = \Metric_\mu^\x(\z) + \Metric_\sigma^\x(\z) + \Metric_\mu^\q(\z) + \Metric_\kappa^\q(\z) ,
    \end{equation}
    \begin{align}
    \text{with}\quad \Metric_\mu^\x(\z) &= \Jac_{\mu_{\bm{\phi}}}(\z)^{\trsp} \Metric_{\ambient}^\x(\mu_{\bm{\phi}}(\z)) \Jac_{\mu_{\bm{\phi}}}(\z) , \nonumber \\  
    \Metric_\sigma^\x(\z) &= \Jac_{\sigma_{\bm{\phi}}}(\z)^{\trsp} \Metric_{\ambient}^\x(\mu_{\bm{\phi}}(\z)) \Jac_{\sigma_{\bm{\phi}}}(\z) , \nonumber \\  
    \Metric_\mu^\q(\z) &= \Jac_{\mu_{\bm{\psi}}}(\z)^{\trsp} \Metric_\ambient^\q(\mu_{\bm{\psi}}(\z)) \Jac_{\mu_{\bm{\psi}}}(\z) , \nonumber \\  
    \Metric_\kappa^\q(\z) &= \Jac_{\kappa{\bm{\psi}}}(\z)^{\trsp} \Metric_\ambient^\q(\mu_{\bm{\psi}}(\z)) \Jac_{\kappa{\bm{\psi}}}(\z) ,  \nonumber
    \end{align}
    where $\Metric_{\ambient}^\x$ and $\Metric_{\ambient}^\q$ represent the position and orientation components of the obstacle-avoidance metric $\Metric_\ambient$, respectively. 
    We emphasize that as the object changes position, the VAE does not need to be re-trained as the change is only in the ambient metric.
    As stated before, obstacles can be avoided only by the end-effector under this task space setting. 

    In the multiple-limb obstacle avoidance setting, rather than just using the end effector, the entire body of the robot is taken into account. This technique takes advantage of redundant solutions in joint space to find a robot configuration that avoids collisions. This is useful in situations where the end effector may be safe from an obstacle, but other parts of the robot's body (such as its links) are in danger of colliding with the obstacle. By providing multiple solutions in joint space during the demonstration phase, the robot can choose a configuration that avoids the obstacle and continue with its task. In cases where the demonstrations do not provide joint space solutions, a variety of obstacle avoidance techniques can be used to move the robot away from obstacles.
    

    There are a variety of methods to incorporate obstacle avoidance into robotic systems. One commonly employed method is to utilize an off-the-shelf obstacle avoidance technique as post-processing of the generated trajectory. This approach involves first generating a trajectory, and then applying an obstacle avoidance technique to prevent collisions with obstacles present in the environment. This method can be relatively straightforward to implement and may effectively avoid obstacles. However, it can also lead to deviation from the intended motion and may not accurately reflect the demonstrations. It is worth noting that obstacle avoidance techniques that operate in joint space are often computationally intensive. Furthermore, this method cannot be applied in the latent space, as the obstacle avoidance technique must have knowledge of geometry and be able to generate trajectories on the manifold.
    

    The technique that will be outlined in the following section offers a capability for efficient and effective obstacle avoidance for multiple limbs, without reliance on additional obstacle avoidance techniques. 

    \subsubsection{Obstacle avoidance in joint space $\R^\eta$:}
\begin{figure}
  \centering
  \includegraphics[width=\linewidth]{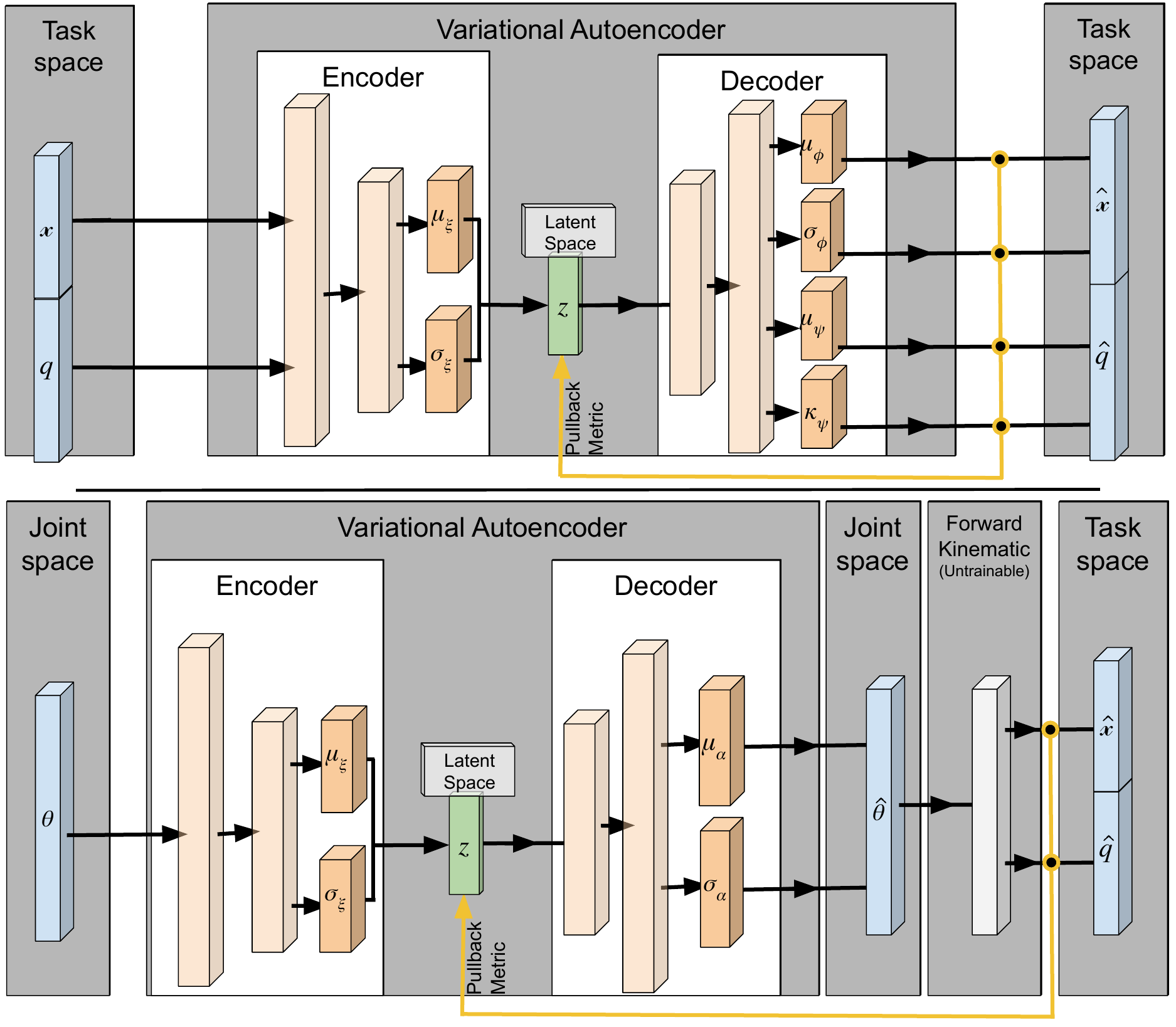}
  \caption{\textit{Top}: The VAE architecture under the task space setting. The blue, orange, green, and gray blocks correspond to ambient spaces, functions with trainable parameters, latent variables, and functions with fixed parameters. The arrows indicate the direction in which the data flows during the query. \textit{Bottom}: The architecture of the VAE under the joint space setting.}
  \label{fig:architecture}
\end{figure}

    Avoiding obstacles at the robot link level while performing motion skills requires considering the whole robot's kinematic structure. 
    Classical motion planning methods model the geometry of the obstacles into the configuration space and later compute an obstacle-free path via sampling methods~\citep{Elbanhawi14:MotionPlanning}.
    In contrast, we take advantage of the forward kinematics layer (see Fig.~\ref{fig:architecture}-bottom), which provides us with task space poses of any point on the robot body, to compute an obstacle-avoidance ambient metric. 
    Similar to the task space formulation presented previously, this ambient metric is then exploited to reshape the learned metric and generate modified geodesic curves that produce collision-free robot movements. 
    
    Specifically, we need to define a collection of points on the robot body $\bm{p}_1, \ldots, \bm{p}_M $ with $\bm{p}_m \in \mathbb{R}^3$. 
    These points are then used to compute the ambient space metric for obstacle-avoidance purposes.
    Therefore, a larger collection of points provides a more robust obstacle-avoidance performance at the cost of higher computational complexity. 
    Given the set of points of interest, we compute an associated ambient metric following Eq.~\eqref{eq:ambient_metric_R3S3} with $\bm{x} = \bm{p}_m$.
    Similar to the task space setting, since the orientation of obstacles is not considered, the corresponding ambient space metric is an identity matrix.
    Finally, we form the whole ambient metric as $\Metric_{\ambient}~=~\operatorname{blockdiag}(\begin{bmatrix}
    \Metric_\ambient^{\bm{p}_1}, \Metric_\ambient^{\bm{p}_2}, \cdots, \Metric_\ambient^{\bm{p}_M}
    \end{bmatrix})$, which is then used to reshape the learned metric of Eq.~\eqref{eq:Joint_space_VAE_Metric} as,
    \begin{equation}
    \Metric(\z) = \Metric^{\bm{\theta}}_\mu(\z) + \Metric^{\bm{\theta}}_\sigma(\z) ,
    \label{eq:Joint_space_VAE_Metric_with_ambient}
    \end{equation}
    with,
    \begin{align*}
        \Metric^{\bm{\theta}}_\mu(\z) &= \left (\Jac_{f_{\textrm{FK}}}(\mu_{\bm{\alpha}}) \Jac_{\mu_{\bm{\alpha}}}(\z) \right ) ^ \trsp \Metric_{\ambient} \left (\Jac_{f_{\textrm{FK}}}(\mu_{\bm{\alpha}}) \Jac_{\mu_{\bm{\alpha}}}(\z) \right ) ,\\
        \Metric^{\bm{\theta}}_\sigma(\z) &= \left (\Jac_{f_{\textrm{FK}}}(\mu_{\bm{\alpha}}) \Jac_{\sigma_{\bm{\alpha}}}(\z) \right ) ^ \trsp \Metric_{\ambient} \left (\Jac_{f_{\textrm{FK}}}(\mu_{\bm{\alpha}}) \Jac_{\sigma_{\bm{\alpha}}}(\z) \right ) .
    \end{align*}

\subsubsection{Generating geodesics on discrete manifolds:}
    
    To generate robot motion, our approach requires computing geodesics, which can be done in several ways \citep{Peyre2010GeodesicMethods}.
    While this can be computed by solving an ordinary differential equation, this approach may not be suitable for real-time robotic applications due to its computational cost, as highlighted in previous studies~\citep{arvanitidis:aistats:2019}.
    A commonly used alternative is to employ gradient descent to minimize the curve length~\citep{Noakes2022FindingGeodesics}.
    While this approach is straightforward and effective in some cases, it may not be optimal when operating in real-time. In our experiments, we found that the gradient descent approach often gets trapped in local minima, depending on a non-trivial initialization. 
    As a consequence, multiple restarts may be required, which increases computational cost. We, therefore, consider an alternative discrete approach.

    As we work with low-dimensional latent spaces, we here propose to simply discretizing them on a regular grid and use a graph-based algorithm for computing shortest paths. 
    If this grid is sufficiently dense, this approach is globally optimal and computationally inexpensive. However, this approach is only feasible in low-dimensional latent spaces due to the curse of dimensionality. In particular, the memory requirements of the graph grow exponentially with dimension, and the short path search grows with the graph size. Our experiments are, however, concerned with low-dimensional latent spaces, where these limitations are negligible.

    In detail, we create a uniform grid over the latent space $\latent$, and assign a weight to each edge in the graph corresponding to the Riemannian distance between neighboring nodes (see Fig.~\ref{fig:graph}). 
    Geodesics are then found using Dijkstra's algorithm~\citep{cormen2009introduction}.
    This algorithm selects a set of graph nodes,
    \begin{align}
      G_\curve = \left \{\g_0, \g_1, \ldots, \g_{S-1}, \g_S \right \}, \quad \g_s \in \R^D, \nonumber
    \end{align}
    where $\g_0$ and $\g_S$ represent the start and the target of the geodesic in the graph, respectively. 
    To select these points, the shortest path on the graph is calculated by minimizing the accumulated weight (cost) of each edge connecting two nodes, computed as in Eq.~\eqref{eq:length_chain_rule}. 
    To ensure a smooth trajectory, we fit a cubic spline $\omega_\lambda$ to the resulting set $G_c$ by minimizing the mean square error.
    The spline computed in $\latent$ is finally used to generate the robot moves through the mean decoder networks: $\mu_{\bm{\theta}}$ and $\mu_{\bm{\psi}}$ or $\mu_{\bm{\alpha}}$. 
    The resulting trajectory can be executed on the robot arm to reproduce the required skill. 

\begin{figure}
        \centering
        \includegraphics[width=\linewidth]{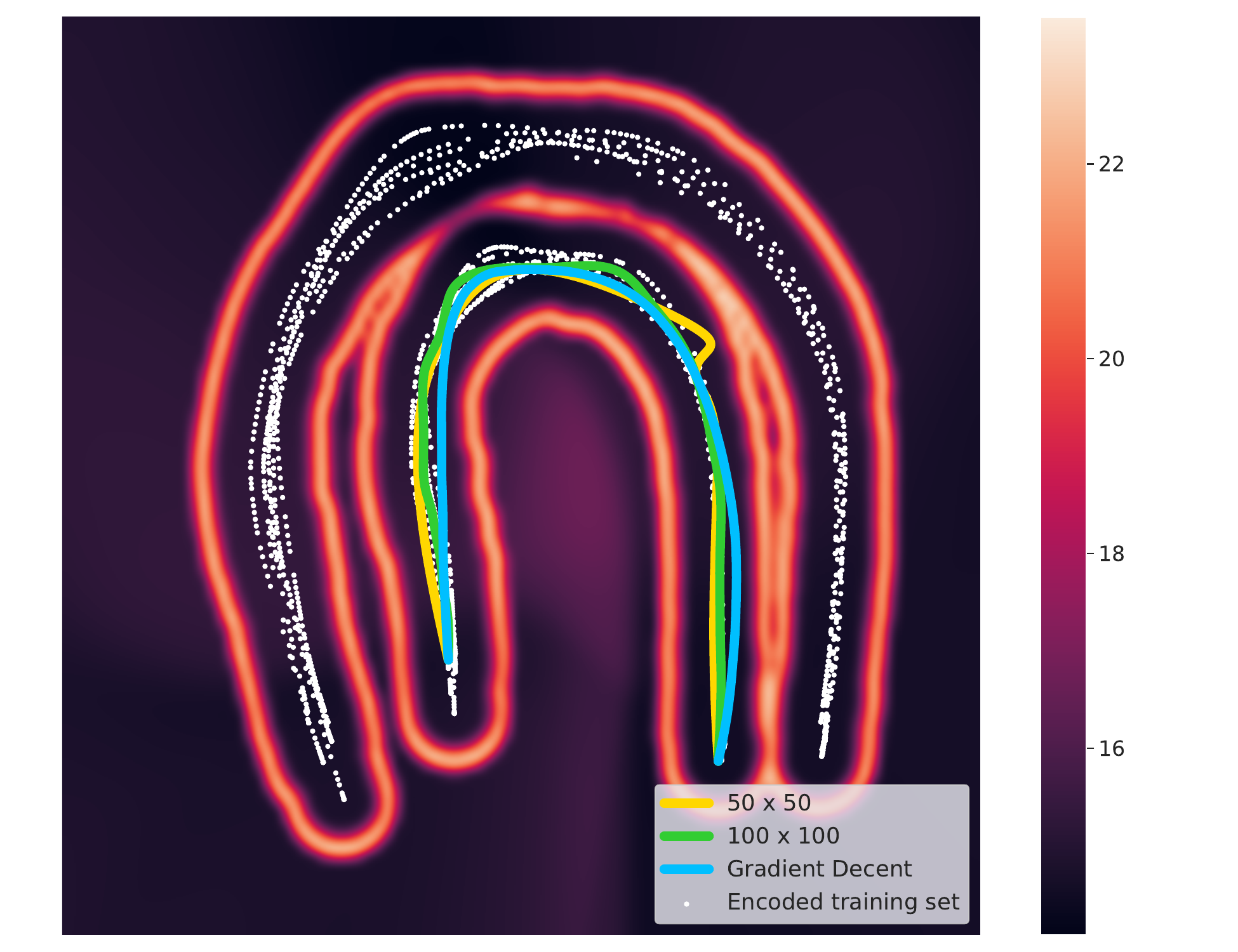}
        \caption{The geodesics generated under continuous and discrete setups. The results show that the geodesic computed using the $50 \times 50$ grid exhibits an unnecessary deviation from the data, likely due to the relatively low resolution of the grid. In contrast, the geodesic generated using the $100 \times 100$ grid has a level of accuracy that is comparable to the (continuous) gradient descent method.}
        \label{fig:Geodesic_Comparison}
    \end{figure}

    In order to verify the validity of the discrete geodesic approximation (graph-based) in comparison to the continuous approximation (gradient-based), a toy example scenario was devised for evaluating their respective accuracies.
    Three different experimental setups were compared, one using gradient descent and two using a graph-based approach. In the graph-based setting, the manifold was discretized, and the Dijkstra algorithm was used to find the shortest path. The discrete setups were also evaluated using two different resolutions to assess the effect of the grid density on the accuracy and efficiency. We employed the same pre-trained model from the toy example presented in Section~\ref{sec:Geodesics_in_task_space}.
    Figure~\ref{fig:Geodesic_Comparison} displays the resulting geodesics. As observed, all methods were successful in approximating a geodesic that follows the trend of the data in the latent space. However, the geodesic generated using the $50 \times 50$ grid, depicted in yellow, exhibits an unnecessary deviation from the data, likely due to the relatively low resolution of the grid. However, the geodesics generated using the $100 \times 100$ grid and gradient descent method exhibit a comparable level of accuracy.

    Moreover, one issue with the graph-based approach is that dynamic obstacles imply that geodesic distances between nodes may change dynamically. 
    To avoid recomputing all edge weights every time an obstacle moves we do as follows.
    Since the learned manifold does not change, we can keep a decoded version of the latent graph in memory (Fig.~\ref{fig:graph}). 
    This way we avoid querying the decoders at run-time. 
    We can then find points on the decoded graph that are near obstacles and rescale their weights to take the ambient metric into account. 
    Once the obstacle moves, we can reset the weights of points that are now sufficiently far from the obstacle.

%% file: Sections/Experiments.tex
\label{sec:experiments}
In this section, we showcase the capabilities of our method in different experiments where human demonstrations are provided in both task and joint spaces. 
We provide a full description of the experimental setup, which includes the design of the VAE networks, Riemannian manifold learning, real-time geodesic computation, and calculation of the ambient metric to avoid obstacles.
We evaluate the performance of our approach using real-robot experiments, namely, \textit{Reach-to-grasp} and \textit{Pouring} tasks. 
The \textit{Pouring} tasks are primarily designed to demonstrate the model's capability to generate multiple-solution trajectories and avoid obstacles under the task space setting. 
The \textit{Reach-to-grasp} task is intended to demonstrate the aforementioned features under a joint space setting. Furthermore, the video is available at: \href{https://sites.google.com/view/motion-generation-on-manifolds/home}{https://sites.google.com/view/motion-generation-on-manifolds/home}.

\subsection{Setup description}
    We consider a set of demonstrations involving a $7$-DOF Franka Emika Panda robot arm endowed with a two-finger gripper. 
    The demonstrations were recorded using kinesthetic teaching at a frequency of $10$Hz. 
    We calculate geodesics on $100 \times 100$ and $50 \times 50 \times 50$ graphs under task space and joint space settings, respectively. All the latent space graphs were bounded by different experimental ranges based on the specific settings of the experiment. For example, our experiments showed that the range of $[-10.0, 10.0]$ is appropriate for the toy example setting for all axes.
    Our Python implementation runs at $100$Hz on ordinary PC hardware. 
    The approach readily runs in real-time. 
    Additionally, obstacles are simulated along with a digital twin of the robot in a simulated environment to provide real-time obstacle information (we did not integrate obstacle localization systems in our setups).

    \begin{figure}
        \centering
        \includegraphics[width=\linewidth]{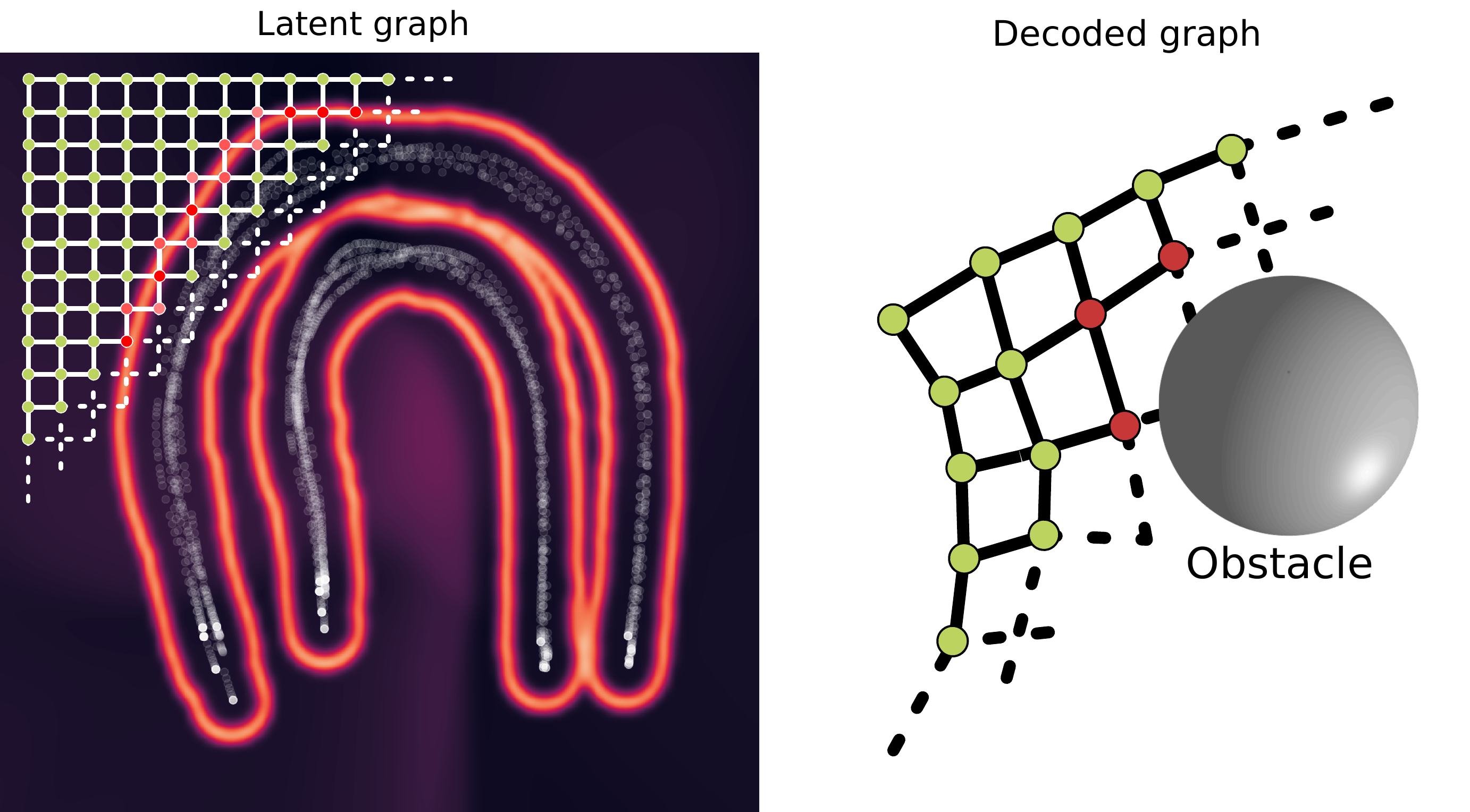}
        \caption{Concept drawing. \emph{Left:} The latent space is discretized to form a grid graph consisting of linearly spaced nodes with edge weights matching Riemannian distances. \emph{Right:} To efficiently handle obstacles, the graph is decoded, such that obstacles can easily be mapped to latent space.}
        \label{fig:graph}
    \end{figure}

\subsection{Architecture}
    In this section, we describe the VAE architecture under both $\R^3 \times \Sph^3$ and $\R^7$ settings. The VAE architectures are implemented using PyTorch~\citep{NEURIPS2019_9015}. 
    
    \subsubsection{VAE architecture in $\R^3 \times \Sph^3$:}
    This particular VAE network is designed to reconstruct end-effector poses in $\R^3~\times~\Sph^3$. 
    The overall architecture is depicted in Figure~\ref{fig:architecture}-\emph{top} with the different components that are required to correctly reproduce end-effector poses. 
    In this architecture, both the decoder and encoder networks have two hidden layers with $200$ and $100$ neuron units (depicted as beige boxes) which output the mean and variance vectors (depicted as orange boxes).  
    Furthermore, as previously explained, the variance and concentration parameters for position and quaternion data are estimated using RBF decoder networks with $500$ kernels calculated by $k$-means over the training dataset~\citep{Arvanitidis:LatentSO} with predefined bandwidth. 
    In this particular setup, the VAE uses a $2$-dimensional latent space (depicted by the green box) to encode the $7$-dimensional input vector (depicted as blue boxes on the left). 
    Moreover, the VAE reconstructs the encoded input back to the ambient space (depicted as blue boxes on the right) using the decoder networks. 
    

        \begin{SCfigure*}[][b]
      \centering
    \centering
    \begin{subfigure}{.78\linewidth}
       \includegraphics[width=1\linewidth]{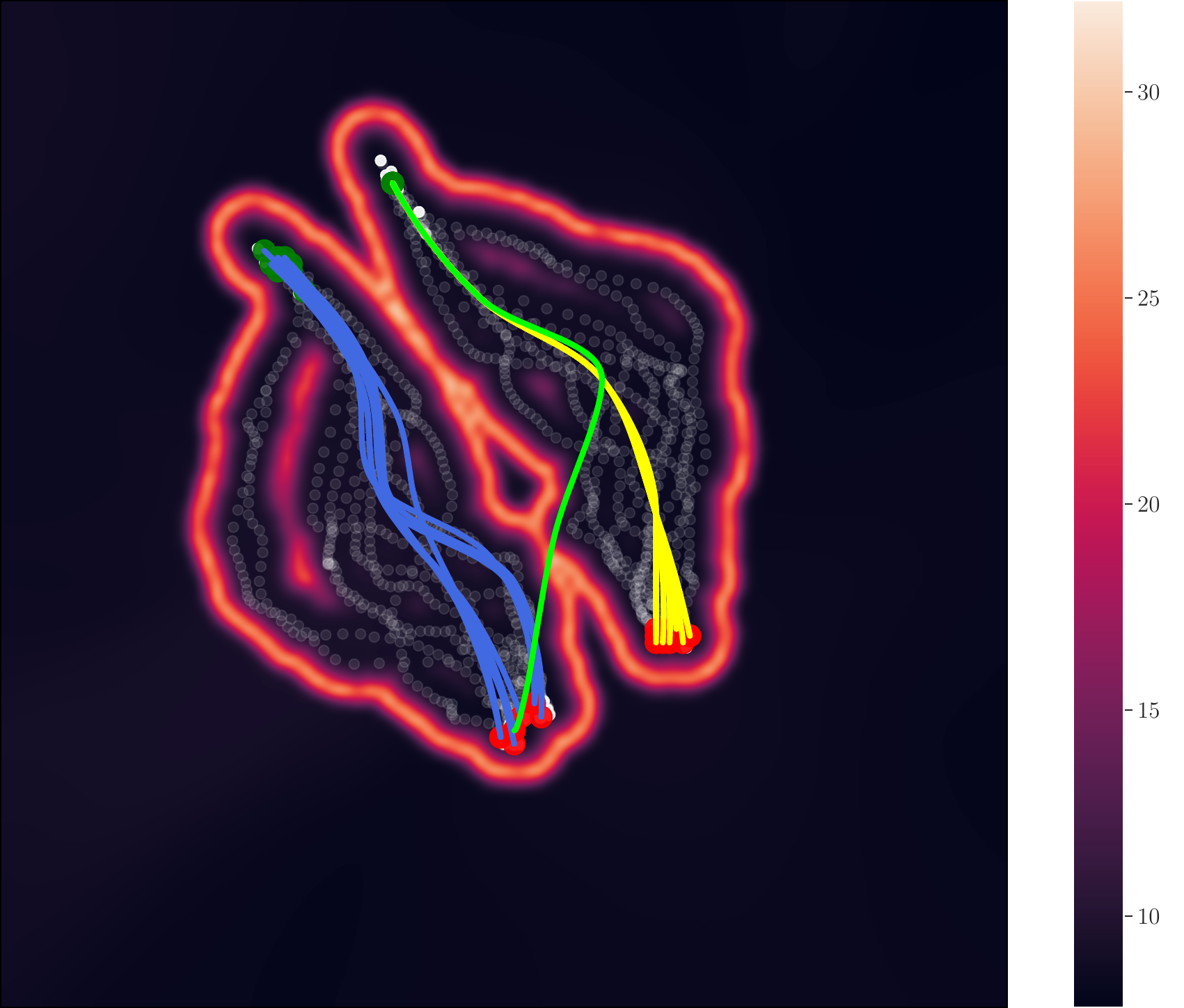}
    \end{subfigure}
    \begin{subfigure}{.7\linewidth}
       \includegraphics[width=1\linewidth]{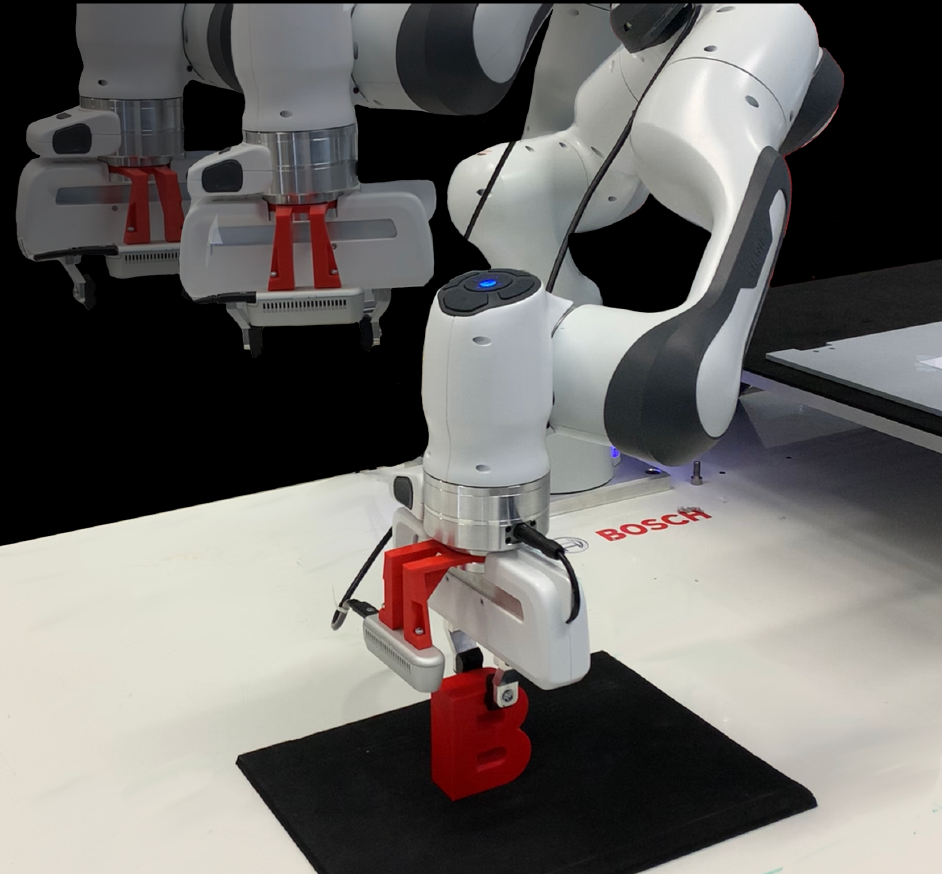}
    \end{subfigure}
      \caption{\emph{Left}: The yellow curves in the right cluster show geodesics starting from the same points and ending up at random targets, and the blue curves in the left cluster connect random points on the manifold. The background depicts the magnification factor derived from the learned manifold, and the semi-transparent white points show the encoded training dataset. \emph{Right}: The decoded geodesic employed on the robot.}
    \label{fig:Grasping_geodesic_task_mf}
    \end{SCfigure*}
    
    \begin{figure}
  \centering
  \includegraphics[width=0.9\linewidth]{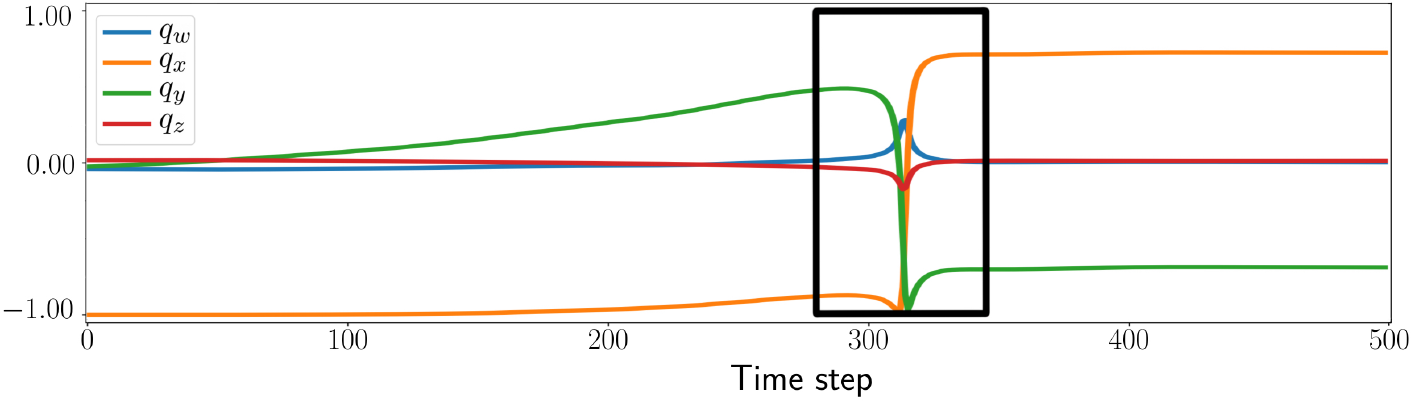}
  \caption{ The evolution of the individual quaternion components of the green geodesic as it passes through high-energy regions.}
  \label{fig:quaternion_flip}
\end{figure}
    
    In this setting, the VAE is implemented using a single neural network as the decoder mean for the position and orientation, while the variance and concentration networks are implemented separately. 
    As a result, the learned metric is defined as,
    \begin{equation}
        \label{eq:obstacle_metric_exp}
        \Metric(\z) = \Metric_{\mu}^{\x,\q}(\z) + \Metric_\sigma^\x(\z) + \Metric_\kappa^\q(\z) , 
    \end{equation}
    \begin{align}
        \text{with} \quad \Metric_{\mu}^{\x,\q}(\z) &= \Jac_{\mu_{\bm{\phi}}}(\z)^{\trsp} \Metric_{\ambient\Q}(\z)\Jac_{\mu_{\bm{\phi}}}(\z) , \nonumber\\
        \Metric_{\ambient\Q}(\z) &= 
        \begin{bmatrix}
    \Metric_\ambient(\mu_{\bm{\phi}}(\z)) & \bm{0} \\  
    \bm{0} & \Metric_\Q(\mu_{\bm{\psi}}(\z))
    \end{bmatrix} , \nonumber\\
        \Metric_\sigma^\x(\z) &= \Jac_{\sigma_{\bm{\phi}}}(\z)^{\trsp} \Metric_\ambient(\mu^\x_{\bm{\phi}}(\z)) \Jac_{\sigma_{\bm{\phi}}}(\z) , \nonumber\\
        \Metric_\kappa^\q(\z) &= \Jac_{\kappa{\bm{\psi}}}(\z)^{\trsp} \Metric_\Q(\mu_{\bm{\psi}}(\z)) \Jac_{\kappa{\bm{\psi}}}(\z) . \nonumber
    \end{align}
    where $\Jac_{\mu_{\bm{\phi}}} \in \R^{(D_\ambient + D_\Q) \times d}$ is the Jacobian of the joint decoder mean network (position and quaternion), and $\Jac_{\sigma_{\bm{\phi}}}~\in~\R^{D_\ambient \times d}$ and $\Jac_{\kappa_{\bm{\psi}}} \in \R^{D_\Q \times d}$ are the Jacobians of the decoder variance and concentration RBF networks, respectively. 
    Since the position and quaternion share the same decoder mean network, the output vector is split into two parts, accordingly. 
    The quaternion part of the decoder mean is projected to the $\Sph^3$ to then define the corresponding von Mises-Fischer distribution as in Eq.~\eqref{eq:vmf_density}. 
    The yellow arrow in Figure~\ref{fig:architecture}-\emph{top} shows the flow of the data from ambient space back to the latent space in order to compute the pullback metric. 
    
    The ELBO parameters $\beta_1$ and $\beta_2$ in Eq.~\eqref{eq:final_elbo} are found experimentally to guarantee good reconstruction of both position and quaternion data.
    It is worth pointing out that we manually provided antipodal quaternions during training, which leads to better latent space structures and reconstruction accuracy.
    The same architecture is used for all the experiments.

    \subsubsection{VAE architecture in $\R^\eta$:}
    Here, we describe our VAE network that reconstructs joint space movements in $\R^7$. 
    The overall architecture is depicted in Figure~\ref{fig:architecture}--\emph{bottom} with different components.
    The input vector (depicted as the blue box on the left) is a joint-value vector representing a single configuration of the robot arm on a trajectory. 
    This vector is fed to the encoder network with two hidden layers of $200$ and $100$ neuron units (depicted as beige boxes) which are the mean and variance vectors for the latent variable (depicted as orange boxes). 
    Moreover, the variance RBF decoder network uses $500$ kernels calculated by $k$-means over the training with predefined bandwidth.

    Under the joint space setting, the VAE uses a $3$-dimensional latent space $\latent$ to encode the input vectors $\bm{\theta}$. 
    As usual, the decoder network reconstructs the encoded inputs back to the joint space. 
    However, in order to access the task space information, necessary for whole-body obstacle avoidance, our architecture is integrated with a forward kinematics layer (depicted as the gray box). 
    Note that this layer is predefined based on the robot arm kinematic model and does not change during training. 
    We leverage this layer to compute task space information regarding multiple points on the robot (and not just the end-effector) given the input configuration vector $\bm{\theta}$.
    To implement this component we used the Python Kinematic and Dynamic Library PyKDL~\citep{PyKDL}.
    
    It is worth noting that as this VAE architecture uses the forward kinematics layer during training, singular kinematic configurations need special attention. 
    The main problem arises in the formulation of our ELBO in Eq.~\eqref{equ:final_ELBO_joint}, which uses the volume measure computed as a function of the determinant of the Jacobian of the forward kinematics function. 
    We can detect singularities when $\det(\Jac_{f_{\textrm{FK}}}(\mu_{\alpha}(\bm{z}_i)))=0$, which may occur due to random initialization of the VAE.
    To circumvent this issue and guarantee that the learning process is not disrupted, we simply add a small regularization term to the kinematic Jacobian.
    
    Finally, we evaluate the obstacle avoidance capabilities in different scenarios where the obstacles partially or entirely obstruct the solutions in joint space. 
    In these experiments, the ambient metric of Eq.\eqref{eq:Joint_space_VAE_Metric_with_ambient} is formulated by considering all the joints on the robot, in addition to the end-effector. 
    This ensures that the robot will avoid obstacles as long as they obstruct the solution for one or more joints or the end-effector. 
    In other words, we do not consider points on the robot links lying between joints, since it was not necessary in our experiments. 
    However, extra points on the robot links can be easily added to guarantee a more robust obstacle avoidance using the whole robot body.

    
    

    \begin{SCfigure*}[][b]
      \centering
    \centering
    \begin{subfigure}{.8\linewidth}
       \includegraphics[width=1\linewidth]{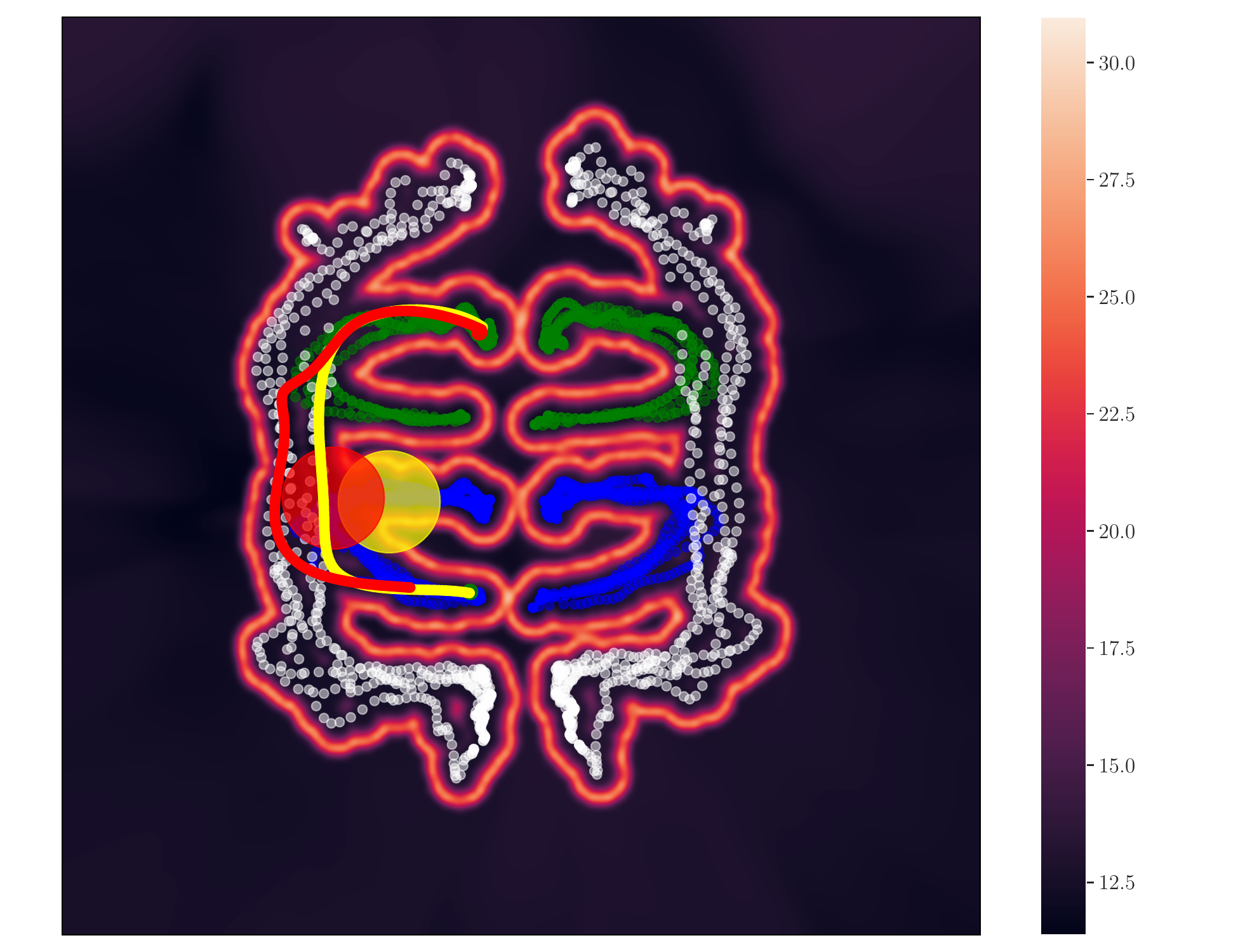}
    \end{subfigure}
    \begin{subfigure}{.6\linewidth}
       \includegraphics[width=1\linewidth]{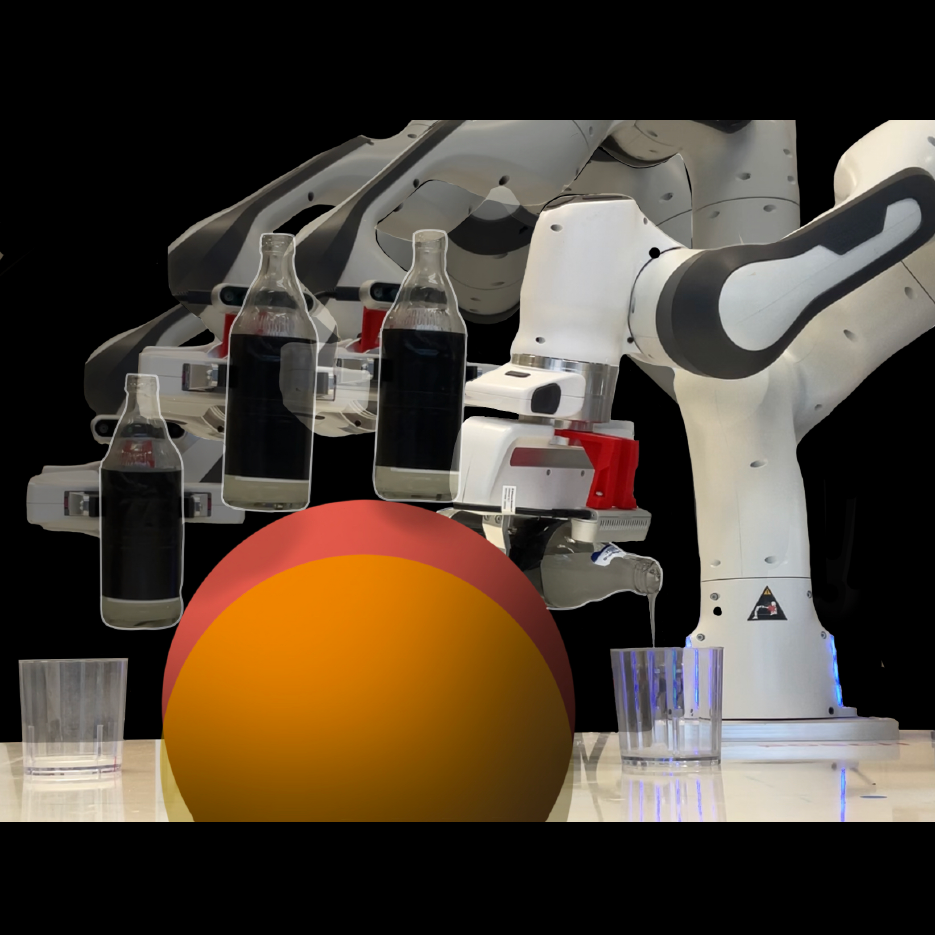}
    \end{subfigure}
      \caption{\textit{Left}:  Red and yellow curves depict geodesics at two different time frames performing dynamic obstacle avoidance with obstacles depicted as red and yellow circles.
      These geodesics also take advantage of a different group of demonstrations to generate a hybrid solution.
      The background depicts the magnification factor derived from the learned manifold with encoded demonstration sets depicted as dot clusters in white, green, and blue. \textit{Right}: The decoded geodesics performed on the robot in different time steps. }
    \label{fig:obstacle_avoidance_pouring_task_mf}
    \end{SCfigure*}

\subsection{Experiments in $\R^3 \times \Sph^3$}
    The first set of experiments focuses on tasks where only the robot end-effector motion is relevant for the task, therefore the demonstrations are recorded in $\R^3 \times \Sph^3$. 

\subsubsection{Reach-to-grasp:}
\label{subsec:Reach-to-grasp}
    The first set of experiments is based on a dataset collected while an operator performs kinesthetic demonstrations of a grasping skill. 
    The grasping motion includes a $90$\textdegree~rotation when approaching the object for performing a side grasp (see Fig.~\ref{fig:Grasping_geodesic_task_mf}).
    The demonstration trajectories start from the same end-effector pose, and they reach the same target position in the task space. 
    To reproduce this grasping skill, we computed a geodesic in $\latent$ which is decoded to generate a continuous trajectory in $\ambient$, which closely reproduces the rotation pattern observed during demonstrations. 
    Figure~\ref{fig:Grasping_geodesic_task_mf}--\emph{left} depicts the magnification factor related to the learned manifold. 
    The semi-transparent white points correspond to the latent representation of the training set, and the yellow and blue curves are geodesics between points assigned from the start and endpoints of the demonstrations. 
    The left panel in Fig.~\ref{fig:Grasping_geodesic_task_mf} shows geodesics in two different scenarios: The yellow geodesics in the right cluster start from the same pose and end up at different targets, while the blue geodesics in the left cluster start and end in different random targets. 
    The results show that the method can successfully generate geodesics that respect the geometry of the manifold learned from the demonstration.
    
    As expected, the magnification factor (Fig.~\ref{fig:Grasping_geodesic_task_mf}--\emph{left}) shows that the learned manifold is composed of two similar clusters, similar to the illustrative example in Fig.~\ref{fig:Toy_example}. 
    We observed that this behavior emerged due to the antipodal encoding of the quaternions, where each cluster represents one side of the hyper-sphere. 
    To confirm this, we generated a new geodesic, depicted in green in the right panel of Fig.~\ref{fig:Grasping_geodesic_task_mf}, which is designed to cross the boundaries of the cluster (start and end locations belong to different clusters). 
    Figure~\ref{fig:quaternion_flip} depicts the evolution of the quaternion components corresponding to the decoded green geodesic.
    As this geodesic curve crosses the clusters, the sign of the end-effector quaternion flips (highlighted by the black rectangle). 
    It is worth emphasizing that by staying on the manifold and avoiding these boundaries, no post-processing of raw quaternion data is required during training or reconstruction. 
    This can be easily guaranteed by monitoring the energy along the curves, which indicates when geodesics approach a high-energy region.
    For instance, the average energy of the blue and yellow geodesics are $7.50$ and $10.51$, respectively, while that of the green geodesic is $2.49 \times 10^9$. 
    As a result, we can simply identify and avoid these geodesics. 
    
    Figure~\ref{fig:Grasping_geodesic_task_mf}--\emph{right} shows the reconstructed geodesic executed by the robot, where the overlapping images display the time evolution of the skill.
    It is easy to observe that the desired motion is successfully generated by our model. 
    Note how the end-effector orientation evolves on the decoded geodesic in the ambient space, showing that the $90$\textdegree~rotation is properly encoded and reproduced using our approach. 

\subsubsection{Pouring:}
    To evaluate our model on a more complicated scenario, we collected a dataset of pouring task demonstrations. 
    The task involves grasping $3$ bottles from $3$ different positions and then pouring $3$ cups placed at $3$ different locations. 
    The demonstrated trajectories cross each other, therefore providing a multiple-solution setting. 
    As a result, with $3$ sets of demonstrations, all $9$ permutations for grasping any bottle from the table and then pouring at any cup are feasible, despite only a small subset of them being demonstrated.
        \begin{SCfigure*}[][!b]
      \centering
    \centering
    \begin{subfigure}{.8\linewidth}
       \includegraphics[width=1\linewidth]{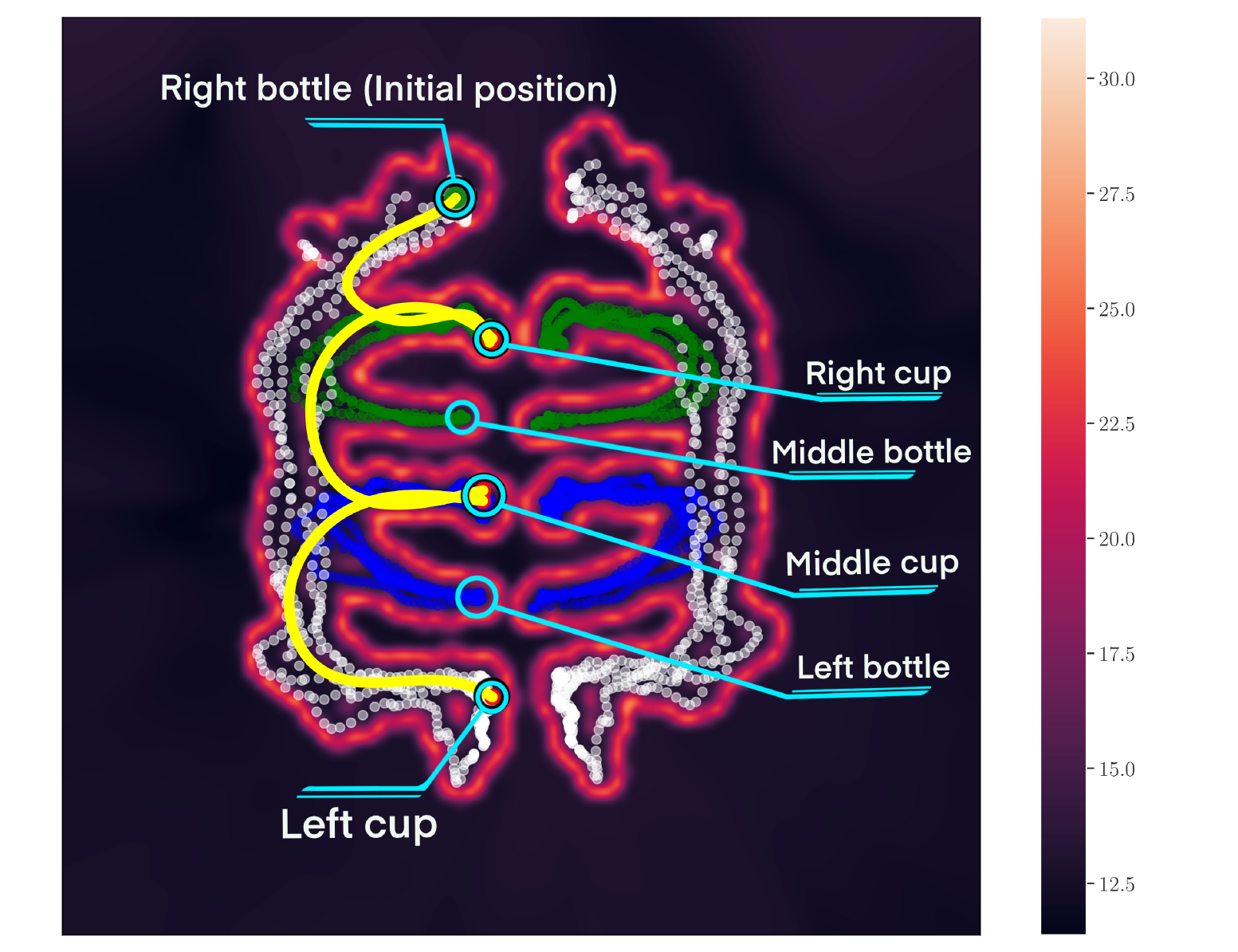}
    \end{subfigure}
    \begin{subfigure}{.6\linewidth}
       \includegraphics[width=1\linewidth]{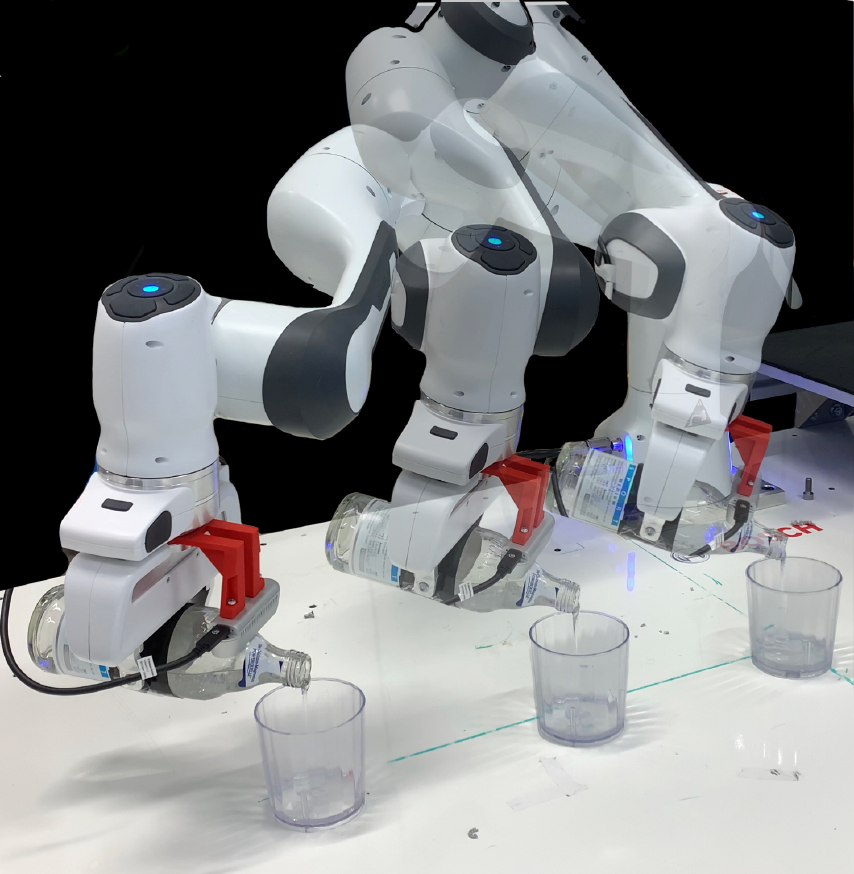}
    \end{subfigure}
      \caption{\textit{Left}: The geodesic shown as the yellow curve combines the blue, green, and white demonstration groups to form a hybrid solution. This experiment uses three geodesics to pour all the cups using a single bottle. \textit{Right}: The decoded geodesics performed on the robot are depicted by superimposing images from different time frames. The transparent robot arms depict the trace of the motion trajectory, which begins with pouring the right cup, the  middle, and lastly the left cup. }
        \label{fig:Multiple_solution_task}
    \end{SCfigure*}

\begin{figure}
  \centering
  \includegraphics[width=\linewidth]{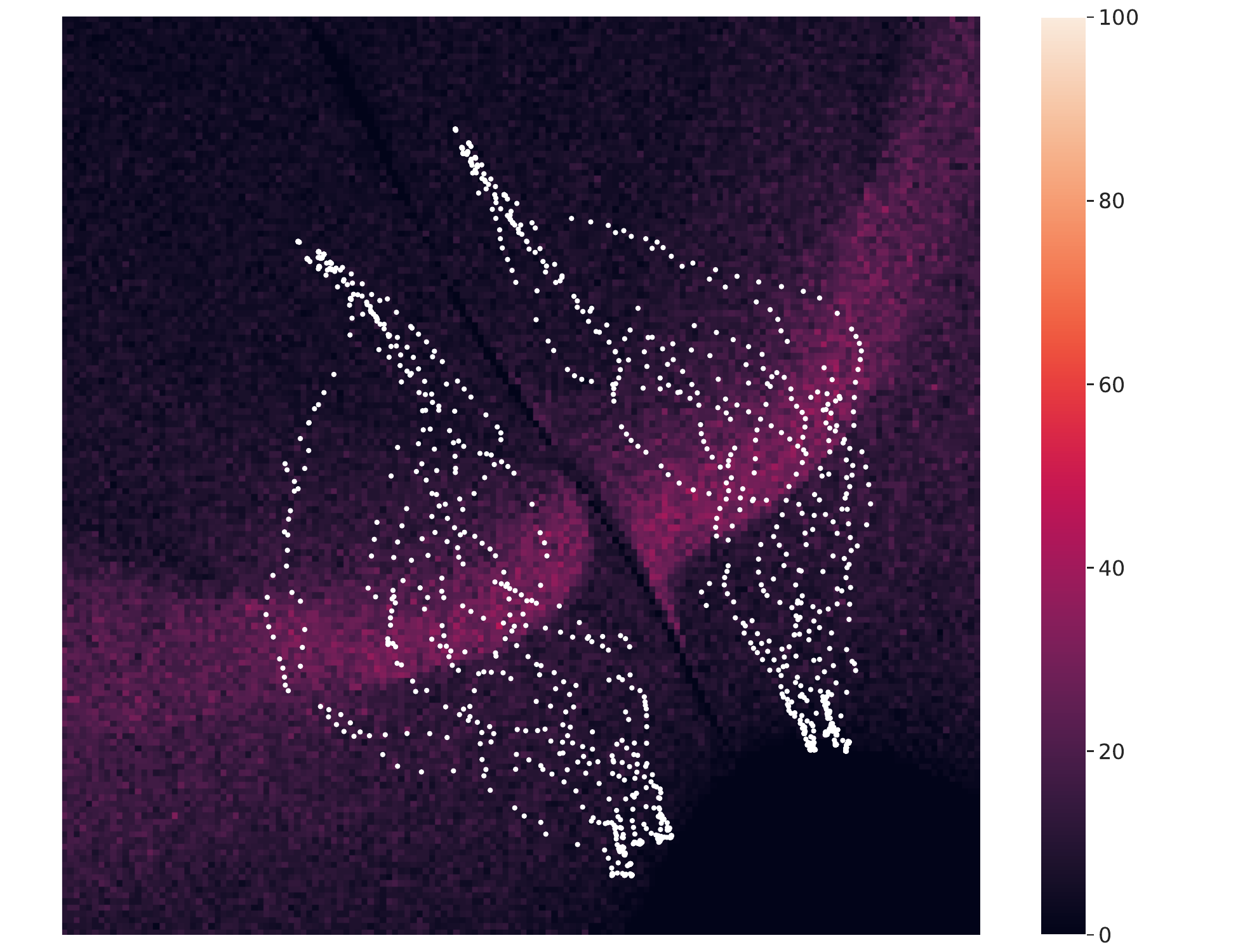}
  \caption{The background illustrates the frequency of failures by an obstacle avoidance technique to provide obstacle-free trajectories for the entire robot body. The white points depict the encoded demonstrations. The results indicate that the robot configurations generated using inverse kinematics in the regions close to the demonstrations are likely to result in collisions with some regions exceeding 90 collisions in 100 trials.} 
  \label{fig:Multiple_limb_Euclidean}
\end{figure}
    
    \begin{SCfigure*}[][t]
      \centering
    \centering
    \begin{subfigure}{.8\linewidth}
       \includegraphics[width=1\linewidth]{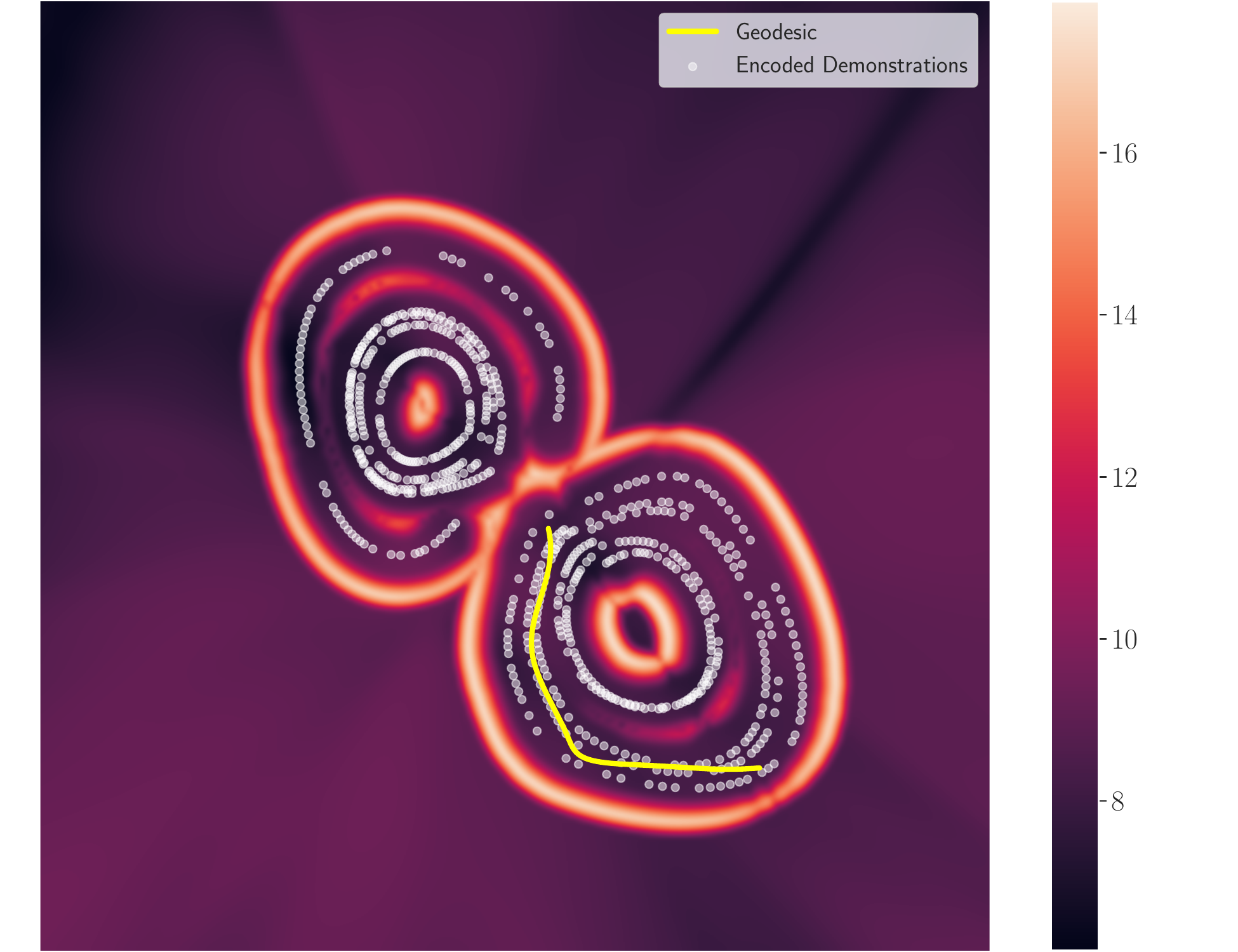}
    \end{subfigure}
    \begin{subfigure}{.7\linewidth}
       \includegraphics[width=1\linewidth]{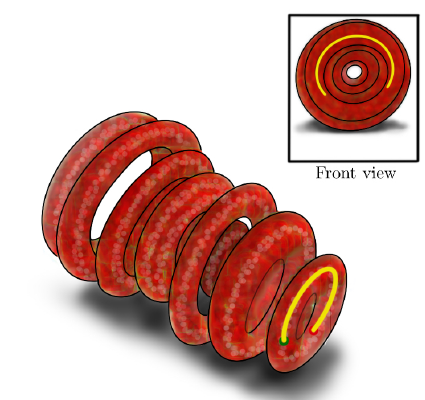}
    \end{subfigure}
      \caption{\textit{Left}: The geodesics calculated in $2$--dimensional latent space, depicted as the yellow curve, reveal several unnecessary transitions between different solutions when connecting two points in the same demonstration.  \textit{Right}: Geodesics computed in $3$--dimensional latent space, shown by the yellow curve, shows that the geodesic does not switch between clusters. }
        \label{fig:geodesic_switch}
    \end{SCfigure*}

    The first feature we want to test in this setting is the obstacle avoidance capabilities via metric reshaping. 
    To do so, we compute the ambient space metric in Eq.~\eqref{eq:ambient_metric_R3S3} based on a spherical obstacle that partially blocks the low-energy regions that the geodesics exploit to find a solution. 
    This way the geodesics are forced to either use the low-energy regions that the individual demonstrations provide or improvise and find a hybrid novel path based on a subset of demonstrations.  
    Figure~\ref{fig:obstacle_avoidance_pouring_task_mf}--\emph{left} shows the geodesic performing obstacle avoidance around a moving obstacle while following the geometry of the manifold.
    Two time instances of the obstacle are depicted as red and yellow circles in the latent space for illustration purposes.
    The red and yellow curves represent geodesics avoiding the red and yellow obstacles, respectively. 
    These curves correspond to one time frame of the adapted geodesic, showing how our method can deal with dynamic obstacles. Fig.~\ref{fig:obstacle_avoidance_pouring_task_mf}--\emph{right} shows the decoded geodesics performed on the robot, where transparent robot arms show the temporal evolution of the skill. 
    In order to correctly perform obstacle avoidance, the parameter $\bm{x}$ in ambient space metric in Eq.~\eqref{eq:ambient_metric_R3S3} represents the position of the bottle when grasped and the obstacle radius $r$ is modified to account for the bottle. To do so, we simplify the bottle geometry by approximating it using a sphere and add its radius $r_{\textrm{bot}}$ to the radius of the obstacle as $r = r_{\textrm{obs}} + r_{\textrm{bot}}$, where $r_{\textrm{obs}}$ is the obstacle radius. 
    This prevents the bottle from colliding with the yellow and red spheres that represent the obstacle. 
    
    Figure~\ref{fig:Multiple_solution_task}--\emph{left} shows the ability to leverage multiple-solution tasks to generate novel movements emerging as combinations of the observed demonstrations.
    Specifically, we generate a combination of three geodesics that can be used to pour all the cups with a single bottle. 
    The geodesic (depicted as a yellow curve) starts from an initial point (depicted in green) in the white demonstrations  group at the top, grasps the bottle, pours the first cup using the green demonstration group, then goes to the second cup using the blue-demonstrations group, and finally pours the third cup using the white-demonstration group again. Figure~\ref{fig:Multiple_solution_task}--\emph{right} uses the same visualization technique to show the decoded geodesics employed on the robot by superimposing images from different time frames. 
    The task begins by pouring the right-side cup, then the center, and lastly the left-side cup.
    As observed, our approach generates the geodesics that properly switch among the demonstrated solutions to reproduce novel movements in ambient space and successfully pour all the cups in the given order.

\begin{figure}
  \centering
  \includegraphics[width=\linewidth]{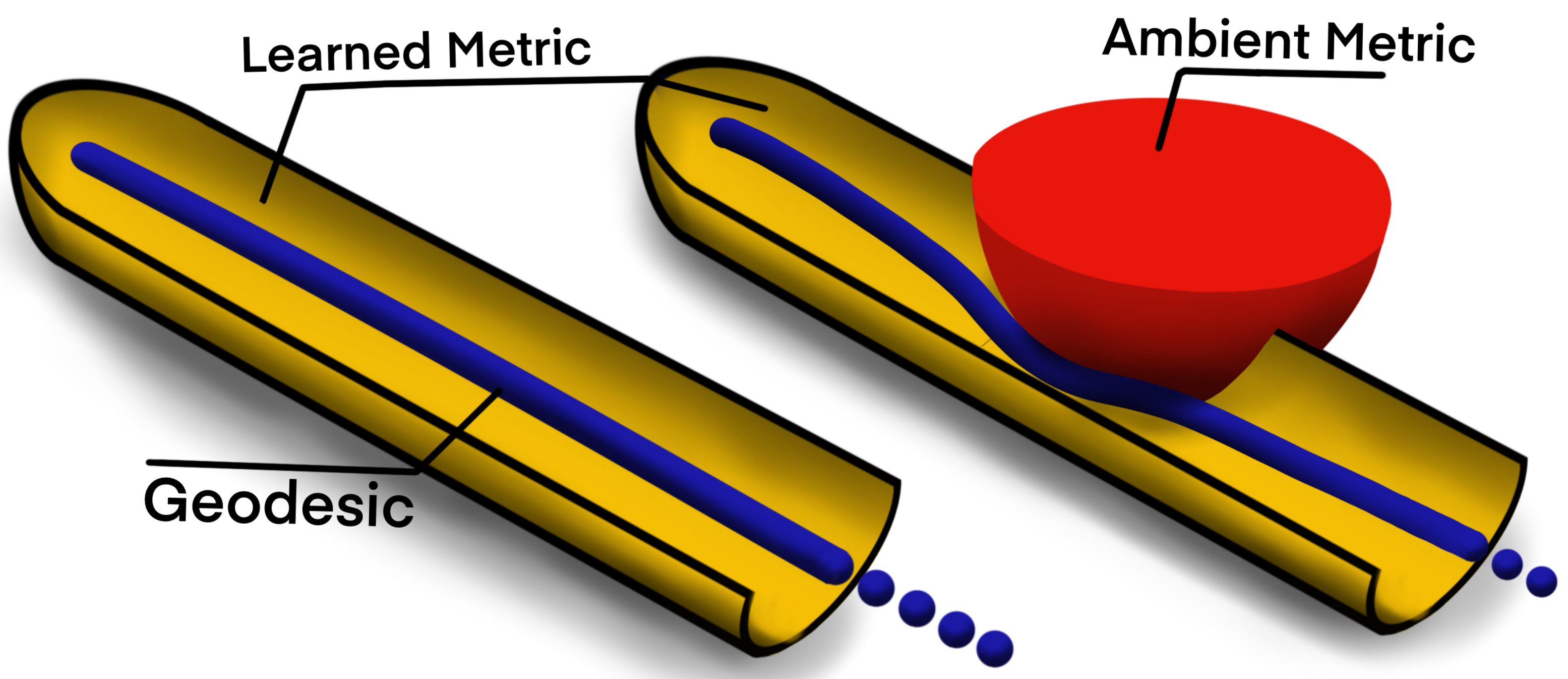}
  \caption{Concept drawing. \textit{Left}: The hollow tube represents the metric, with low-energy regions inside surrounded by high-energy boundaries. The geodesic depicted as the blue curve travels through the low-energy regions. \textit{Right}: The hollow tube is partially blocked by a solid high-energy region corresponding to the ambient space metric representing the obstacle. The geodesic depicted as the blue curve successfully travels through the low-energy regions meanwhile avoiding the obstacle.  }
  \label{fig:3D_metric_concept_drawing}
\end{figure}

\begin{figure*}
  \centering
    \begin{subfigure}{.33\linewidth}
    \centering
    \includegraphics[width = \linewidth]{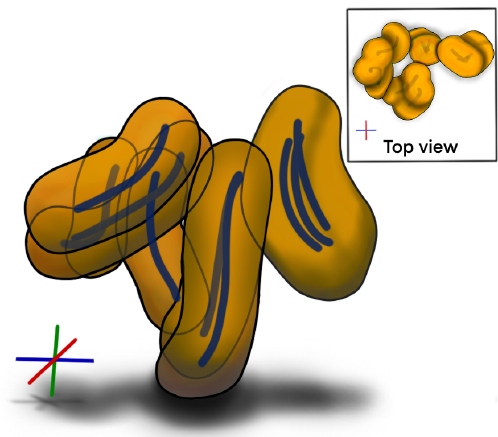}
  \end{subfigure}%
  \begin{subfigure}{.33\linewidth}
    \centering
    \includegraphics[width = \linewidth]{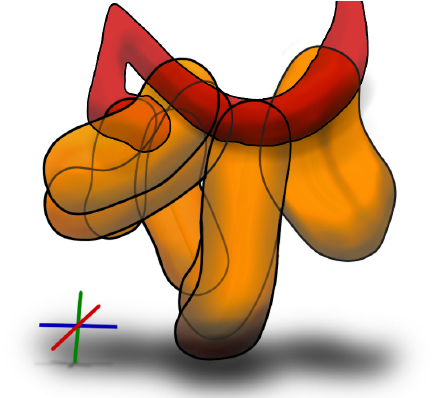}
  \end{subfigure}
    \begin{subfigure}{.33\linewidth}
    \centering
    \includegraphics[width = \linewidth]{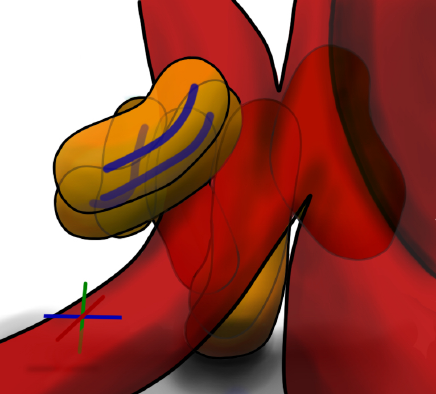}
  \end{subfigure}
  \caption{\emph{left}: The magnification factor in $3$--dimensional latent space contains hollow tubes representing the learned Riemannian metric. These hollow tubes contain low-energy regions surrounded by high-energy boundaries. Several geodesics are calculated and visualized in blue. The manifold's low-energy areas are rendered transparent. \emph{middle}: The same magnification factor when an obstacle is introduced in such a manner that all viable solutions in the ambient space are blocked. Due to the fact that the obstacle introduces a high energy zone (in red) that goes through all of these hollow tubes, none of the geodesics are feasible, therefore no geodesic is shown in this plot.  \emph{right}: The same magnification factor when an obstacle is introduced in such a manner that partially obstructs the solutions in the ambient space. Several geodesics (blue curves) were left outside of the obstacle region on the left side of the panel. }
  \label{fig:Joint_Space_90_Grasp_Separated}
\end{figure*}

    \begin{SCfigure*}[][h]
  \centering
\centering
\begin{subfigure}{.66\linewidth}
   \includegraphics[width=1\linewidth]{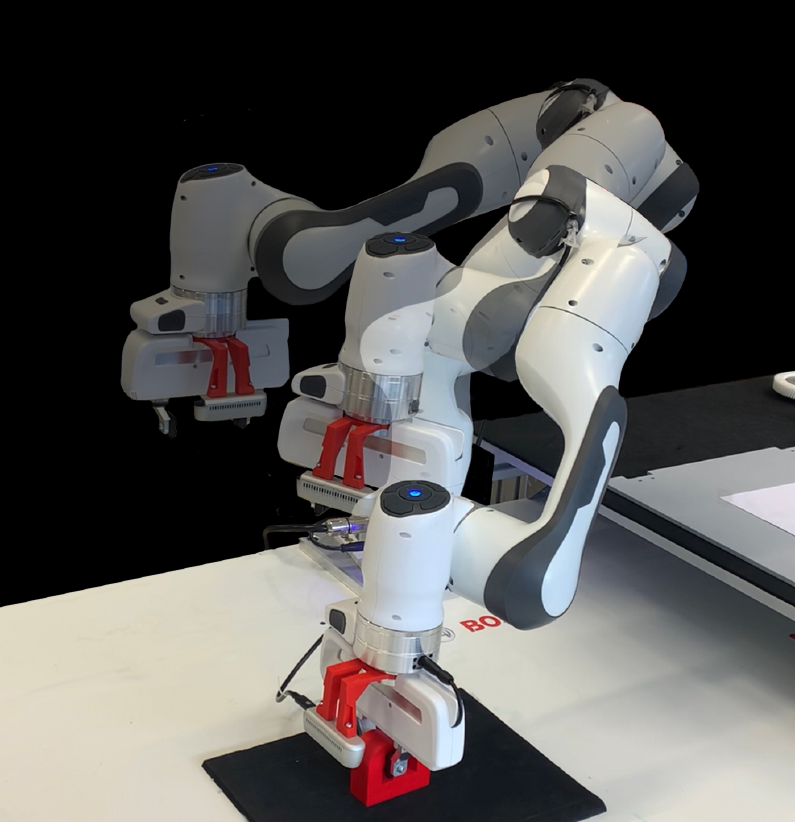}
\end{subfigure}
\begin{subfigure}{0.7\linewidth}
   \includegraphics[width=1\linewidth]{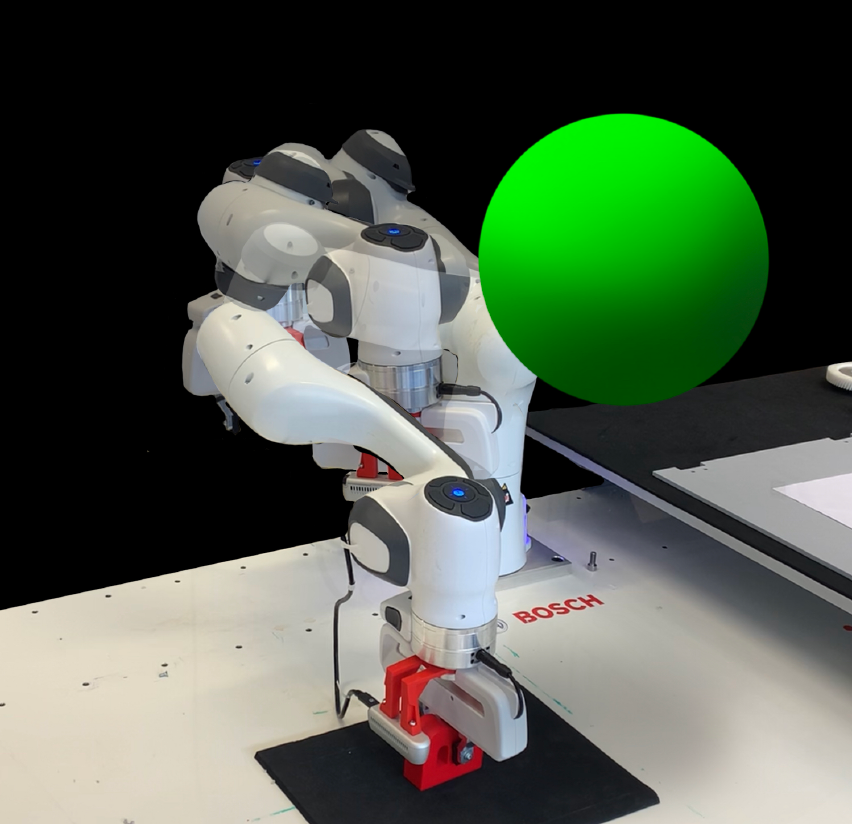}
\end{subfigure}
  \caption{\textit{Left}: The decoded geodesic is employed on the robot arm when no obstacle is present in the environment.  \textit{Right}: The decoded geodesic employed on the robot arm with an obstacle partially obstructing the solutions. The obstacle is depicted as the green sphere.    }
    \label{fig:Joint_Space_90_Dgrasp_Separated_robot}
\end{SCfigure*}

\subsubsection{Multiple-limb obstacle avoidance:}

Multiple-limb obstacle avoidance is a technique that aims to avoid obstacles by considering the entire robot body, rather than just the end-effector. This approach offers a more comprehensive way of obstacle avoidance as compared to techniques that focus exclusively on end-effector motion. To implement this method, we leverage redundant solutions in joint space from demonstrations. Therefore, by operating in the joint space, we can achieve redundancy at the joint level, which is essential for the success of multiple-limb obstacle avoidance. When working under the task space setting, it is not possible to achieve the redundancy in the joint level required for multiple-limb obstacle avoidance. Thus, attempts to avoid obstacles using multiple limbs may be unsuccessful. However, an off-the-shelf obstacle avoidance technique can be used to navigate the robot away from obstacles and complete the task under the task space setting.

To evaluate this, we used the trained model from Sec.~\ref{subsec:Reach-to-grasp} to show how likely an obstacle avoidance technique fails to provide an obstacle-free joint configuration. To do so, we decoded a $150 \times 150$ equidistant latent grid to access their corresponding input space states which represent the end-effector poses. Then, we used each pose to compute $100$ different joint configurations using inverse kinematics. 
The obstacle was placed in a way that was close to the robot body but did not block any demonstration in the task space, as depicted in Fig.~\ref{fig:Joint_Space_90_Grasp_Connected}. Therefore, this setting requires multiple-limb obstacle avoidance in order to successfully perform the task. Each configuration was scored $1.0$ even if a single point on the robot collided with the obstacle, otherwise, the score was $0.0$. In Fig.~\ref{fig:Multiple_limb_Euclidean}, the background shows the sum of these scores for each latent state over the latent grid, and the white points depict the encoded demonstrations in the latent space. The results show that in the regions close to demonstrations, there is up to $90$ percent chance to generate a joint configuration that leads to a collision.

\subsection{Experiments in the joint space $\mathbb{R}^\eta$}
    In this section, we focus on tasks where joint-level motion patterns are relevant, and therefore the human teacher provides kinesthetic demonstrations in $\R^\eta$, where $\eta$ is the number of DOF of the robot. 
    When learning Riemannian metrics in this setting, we initially designed a $2$-dimensional latent space for our VAE, which proved to be insufficient to encode the demonstrated motion patterns. 
    Specifically, we analyze the capacity of the latent space to encode the skill manifold experimentally.
    To do so, we investigated the switching behavior of geodesics in $\latent$.
    In other words, overlap between low-energy regions in $\latent$, representing two or more different demonstration sets, may lead to unnecessary and frequent switches between these solutions when computing a geodesic. To put it differently, when computing a geodesic, switching between two demonstration sets is unavoidable when the total energy of the geodesic switching between them is less than the total energy without the switch.
    However, switching behaviors may be avoided by having high-energy regions among demonstration clusters in $\latent$.
    Furthermore, specifically under the joint space setting, the frequent switching in geodesics may lead to jittery motion in task space when applied to the robot. 
    To illustrate this issue, we designed a simple experiment in which a robotic arm follows a circular pattern with its end-effector.
    The start and target configuration of the geodesic is selected from the same demonstrated trajectory, and the objective is to evaluate if the geodesic stays on the low-energy regions corresponding to the same demonstrated trajectory.

    Fig.~\ref{fig:geodesic_switch}--\emph{left} shows a geodesic curve computed in the $2$-dimensional latent space depicted as the yellow curve. 
    This geodesic exhibits unnecessary switches among different solutions (i.e.\@ circular white demonstrations). 
    Therefore, when the decoded geodesic is deployed on the robot, it results in jerky movements and undesirable back-and-forth motions. 
    To solve this issue, we evaluated the same experiment using a $3$-dimensional latent space. Figure~\ref{fig:geodesic_switch}--\emph{right} shows the magnification factor of the metric learned using the same training data but in a $3$--dimensional latent space. 
    This magnification factor shows several torus-like clusters, representing separate demonstration groups instead of collapsing them into a plane, as in the $2$-dimensional case. 
    Moreover, the resulting geodesic (depicted as the yellow curve) does not switch among solutions, which provides stable and smooth robot end-effector movements when decoded.

    To provide further details on the learned manifold in the $3$-dimensional latent space, we create an illustration shown in Fig.~\ref{fig:3D_metric_concept_drawing}, where the learned metric is shown as yellow hollow tubes. 
    Their inner part encodes low-energy regions which are surrounded by high-energy boundaries.
    Figure~\ref{fig:3D_metric_concept_drawing} displays a horizontal cut to show the inner part of the learned metric. 
    To illustrate the obstacles, the right hollow tube is partially blocked by the red sphere representing the ambient space metric in the latent space $\latent$, which is a solid high-energy region. 
    Additionally, the figure depicts a geodesic curve traveling successfully along both tubes, showcasing a collision-free trajectory at the right-side plot.

\subsubsection{Reach-to-grasp:}
    Similar to the task space setting, we used the reach-to-grasp task to evaluate our approach under the joint space setting. 
    In this case, the demonstrations are quite similar at the end-effector level but differ at the joint space, as we exploited the kinematic robot redundancy to provide different joint trajectories. 
    Figure~\ref{fig:Joint_Space_90_Grasp_Separated}--\emph{left} shows the magnification factor in a $3$-dimensional latent space, where it can be seen that the learned metric corresponds to several connected and separated hollow tubes. 
    As mentioned previously, these hollow tubes have low-energy inner regions surrounded by high-energy boundaries, analogous to the learned metric in a $2$-dimensional latent space. 
    As shown in the figure, the generated geodesics stay inside the tubes and avoid crossing the boundaries. 
    Figure~\ref{fig:Joint_Space_90_Dgrasp_Separated_robot}--\emph{left} shows the robot executing the decoded geodesic by applying the joint values directly on the robot using a joint position controller. 
    We can observe that the decoded geodesic is able to generate the demonstrated grasp motion with $90$\textdegree~rotation during the approaching part. 
    Note that the start and end points of the geodesics are extracted from the demonstrations.

    Figure~\ref{fig:Joint_Space_90_Grasp_Separated}--\emph{middle} displays the magnification factor where an obstacle is placed in such a way that all possible solutions in joint space are blocked (e.g. obstacle placed on the common target of all the demonstrations).
    Note that the obstacle introduces a high energy zone (depicted in red) that passes through all of the hollow tubes representing the learned manifold, therefore, none of the geodesics are feasible. 
    The feasibility of the geodesics is assessed by evaluating the total energy of the curve. If a geodesic passes through an obstacle, its energy will suddenly increase, providing an indication of a potential collision.
    Additionally, Fig.~\ref{fig:Joint_Space_90_Grasp_Separated}--\emph{right} illustrates the same learned metric but reshaped using a different ambient space metric. 
    We can now see that the obstacle partially obstructs the possible solutions, and therefore some few geodesics (depicted as blue curves) are still able to successfully generate obstacle-free movements. Figure~\ref{fig:Joint_Space_90_Dgrasp_Separated_robot}--\emph{right} shows the robot executing the decoded geodesic using a joint position controller while avoiding the obstacle.
    
\begin{SCfigure*}[][t]
  \centering
\centering
\begin{subfigure}{.66\linewidth}
   \includegraphics[width=1\linewidth]{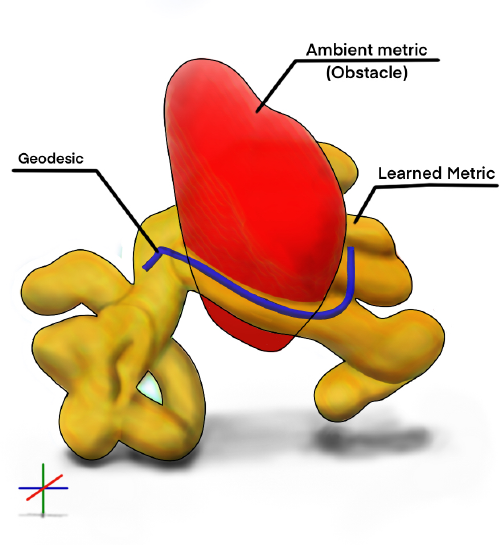}
\end{subfigure}
\begin{subfigure}{.7\linewidth}
   \includegraphics[width=1\linewidth]{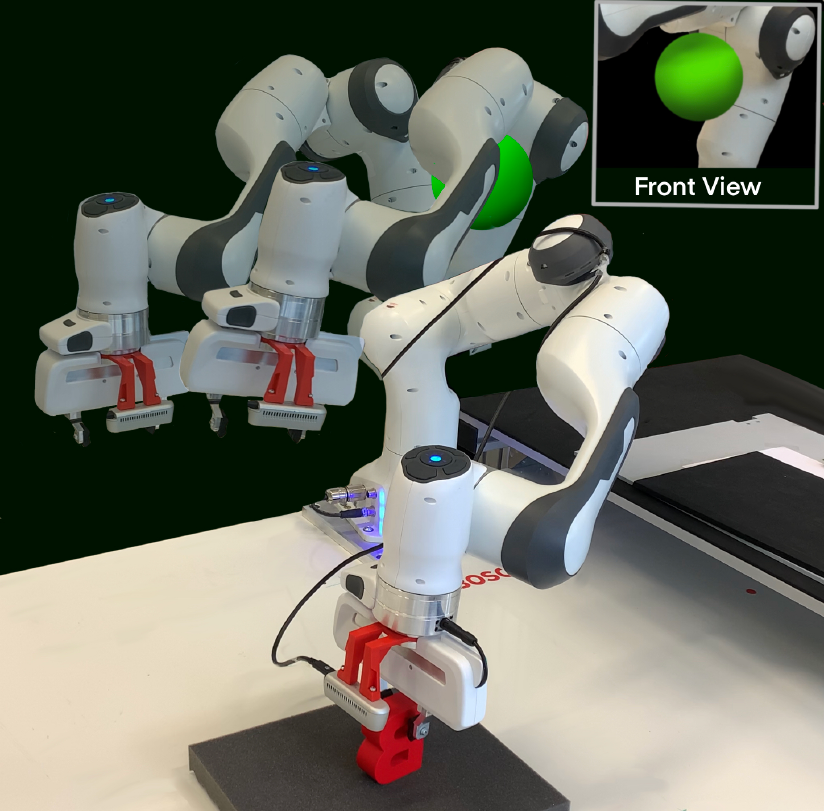}
\end{subfigure}
  \caption{\textit{Left}: The geodesic depicted as the blue curve avoids the obstacle by traveling through different demonstrations. \textit{Right}: The decoded geodesic employed on the robot arm. Images from different time steps are superimposed to show the trace of the motion. 
  The green sphere represents the obstacle in the ambient space. }
    \label{fig:Joint_Space_90_Grasp_Connected}
\end{SCfigure*}

    We designed another experiment to showcase how the multiple-solution capabilities can be leveraged to generate collision-free movements. 
    If the different demonstrated joint space trajectories sufficiently overlap, the learned manifold will be characterized by several overlapping low-energy regions (i.e. hollow tubes), which geodesic curves can travel through. 
    As a result, if the obstacles partially block the learned manifold, the geodesic may still travel among solutions to generate new movements out of combinations of the provided demonstrations.
    To show this behavior, we used a different demonstration dataset where we accounted for several overlapping solutions starting from the same joint configuration. 
    Figure~\ref{fig:Joint_Space_90_Grasp_Connected}--\emph{left} shows the magnification factor of the learned Riemannian metric in the $3$--dimensional latent space. 
    The learned manifold and the ambient metric can be distinguished visually based on their energy values. 
    The ambient metric, which represents the obstacle regions (depicted in red), encodes high-energy zones. 
    The learned manifold (depicted in yellow) is characterized by several entangled hollow tubes.
    The geodesic, shown as the blue curve, begins from the common start configuration on the left and successfully navigates around the obstacle to reach the target on the right side.
    Figure~\ref{fig:Joint_Space_90_Grasp_Connected}--\emph{right} displays the decoded geodesic deployed directly on the robot arm using a joint position controller.
    Note how the robot arm avoids the obstacle by carefully maneuvering around it while successfully performing the grasping skill.

\subsubsection{Pouring:}
\begin{SCfigure*}[][t]
\centering
\centering
\begin{subfigure}{.66\linewidth}
   \includegraphics[width=1\linewidth]{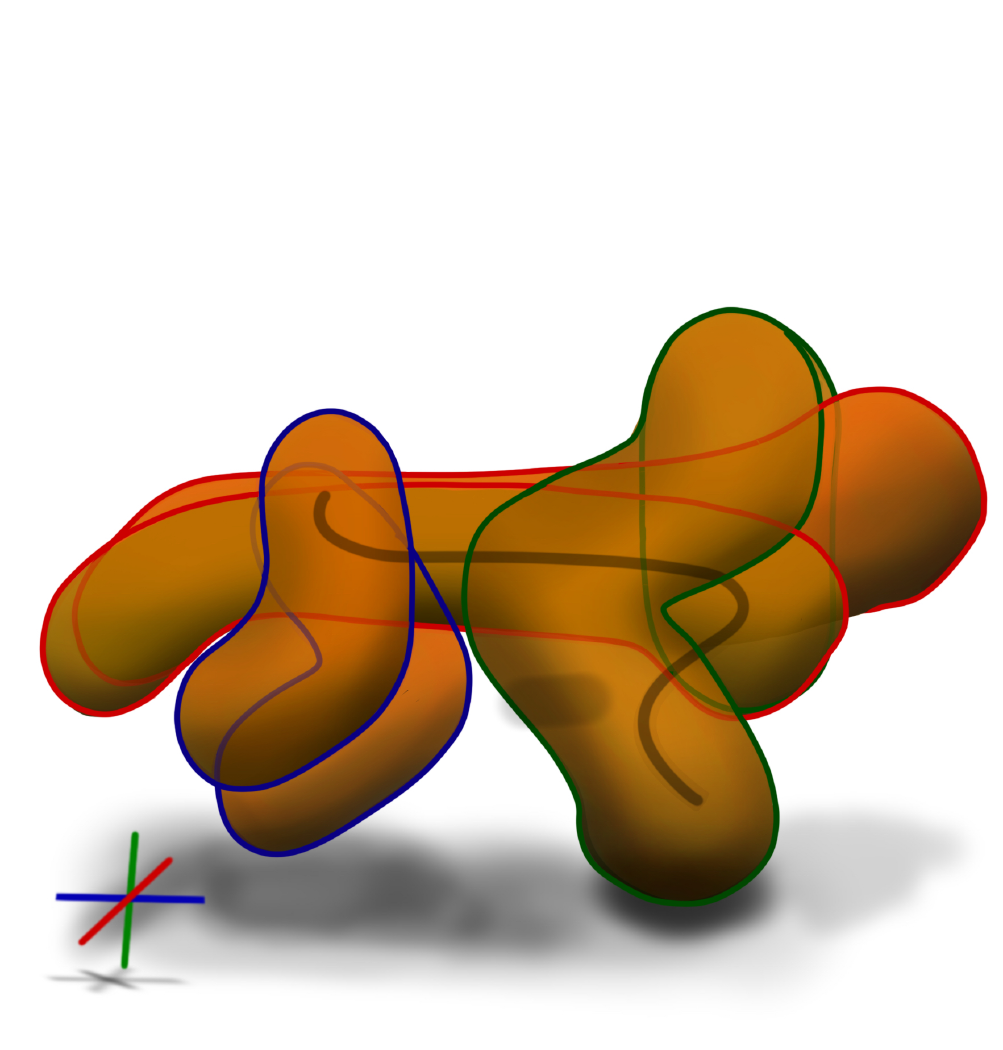}
\end{subfigure}
\begin{subfigure}{.7\linewidth}
   \includegraphics[width=1\linewidth]{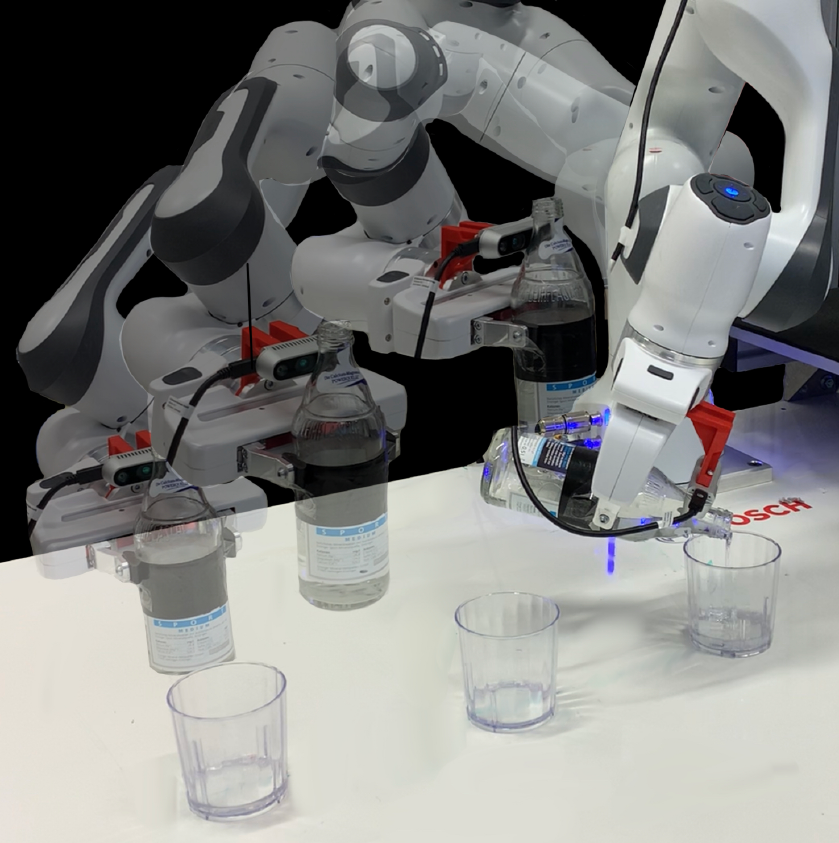}
\end{subfigure}
  \caption{\emph{Left}: The magnification factor in $3$--dimensional latent space representing the learned Riemannian metric. This metric contains low-energy regions surrounded by high-energy boundaries depicted as red hollow tubes. The geodesic curve is depicted in black. The outline color of each tube (green, red, or blue) indicates which demonstration group it belongs to. \emph{Right}: The decoded geodesic employed on the robot. }
    \label{fig:Joint_Space_Pouring_Separated}
\end{SCfigure*}

    To evaluate the method in a more complex scenario, we also performed the pouring task under the joint space setting.
    Similar to the task space experiment, the demonstrations also overlap in joint space. 
    For this specific experiment, we mainly focus on evaluating the multiple-solution trajectories.
    Figure~\ref{fig:Joint_Space_Pouring_Separated}--\emph{left} shows the magnification factor in $3$-dimensional latent space showing entangled hollow tubes, which represent the learned Riemannian metric. 
    Each hollow tube encodes low-energy regions surrounded by high-energy boundaries, the latter outlined by green, red, and blue solid lines. 
    Each color represents one group of demonstrations. 
    Since the magnification factor learned with the original dataset of $5$ demonstrations per group (corresponding to the bottle's initial position) is very hard to visualize, we opt for simplicity and used a subset of $2$ demonstrations per bottle. 
    As shown, the geodesic (depicted as the black curve) starts from a point in the second demonstration group (green border) and switches to the first group (red border) to reach the target located in the third group (blue boundary).
    Figure~\ref{fig:Joint_Space_Pouring_Separated}--\emph{right} shows that the decoded geodesic employed on the robot successfully performs the pouring task. 
    It is also evident the geodesic uses a multiple-solution strategy to reproduce a new trajectory that was not explicitly demonstrated to the robot in the training phase.


%% file: Sections/Discussion.tex
\label{sec:discussion}
We presented a novel LfD approach that learns Riemannian manifolds from human demonstrations, over which geodesic curves are computed and used as a motion generator. 
These geodesics are capable of recovering and synthesizing robot movements similar to the provided demonstrations. 
Therefore, our technique enables robot motion planning between two arbitrary points on the learned manifold via geodesics. 
This is achieved through a variational autoencoder (VAE) from which a pullback Riemannian metric can be computed in the VAE latent space and used to generate movements in both task and joint spaces. 
Additionally, we proposed reshaping the learned Riemannian metric using ambient metrics to avoid static or dynamic obstacles on the fly. 
This metric reshaping along with a forward kinematics layer enabled multiple-limb obstacle avoidance capabilities, which allowed us to achieve reactive motions. 
In order to guarantee fast geodesic computation and motion generation, we developed a graph-based technique built on the Dijkstra algorithm to provide real-time motion synthesis and adaptation. 
We extensively tested our approach in both joint and task space settings, showing that our geodesic motion generation performs effectively in both settings and in a variety of tasks such as grasping and pouring.
We would like to draw attention to the fact that the developed system has certain limitations and potential areas for improvement. In this regard, we would like to present a selection of such aspects for discussion.

\subsubsection*{Geodesics generalization:}
During our experiments, we observed that when the geodesic curves are forced to leave the learned manifold, thus crossing high-energy boundaries, the parts of the geodesic that lie outside the manifold may cause inaccurate and undesirable motions. 
This is a potential issue when the given start or target points are placed outside the data manifold, or when the obstacles fully blocked it, which forces the geodesics to leave the manifold to comply with the desired task specifications.
We believe this is a consequence of using VAEs to learn the skill manifold where the data lying outside the training data support may be arbitrarily misrepresented in the latent space $\latent$. 
We hypothesize that this problem may be addressed by learning a bijective mapping between old demonstrations and new conditions, and then using this function to transform the learned manifold (e.g., by expanding or rotating) to fit another region of the space.

\subsubsection*{Obstacle avoidance:}
Regarding our obstacle avoidance approach, the ambient metric used to reshape the learned metric enforces a ``soft constraint" rather than a ``hard constraint". 
This means, under certain situations, the geodesic might still cross the obstacle instead of avoiding it. 
It is worth noting that we were unable to replicate this scenario since the generated geodesics tended to avoid the obstacle by abandoning the manifold rather than crossing the obstacle, which can be easily detected and prevented if necessary.
To address this potential issue, the graph-based geodesic can be further exploited by removing the nodes near the obstacle from the graph instead of re-weighting the edges. 
While this strategy may reduce the computational overhead and work in practice, it is not theoretically grounded.
Finally, our obstacle avoidance formulation only considered simple obstacles, but the strategy can be extended to multiple dynamic obstacles. 
Instead of working with single balls, one can imagine extending the approach to complex obstacle shapes represented as point clouds. 
This may increase the implementation demands in order to remain real-time, but such an extension seems reasonable. 

\subsubsection*{Latent space topology:}
As explained in Section~\ref{sec:experiments}, we observed that the latent space dimensionality plays a critical role when learning joint space motions.
As reported, a $3$-dimensional latent space was necessary to calculate smooth geodesic curves for joint space tasks. 
This resulted in increased computational complexity when calculating the ambient metric and geodesics in the latent space.
This implies that this issue may exacerbate when working with high-DOF robots, requiring higher-dimensional latent spaces, and therefore raising the need to design more efficient geodesic computation methods.
Note that this phenomenon may be related to the latent space topology, which is here assumed to be Euclidean in accordance with the used VAE Gaussian prior.
We plan to theoretically and experimentally analyze the effects of the latent space topology when learning Riemannian manifolds for robot motion generation. 

\subsubsection*{Scalability to high-dimensional latent spaces:}
In our current setup, we used task- and joint-space trajectories, and therefore the latent space did not encode any external perceptual information. In this setup, low-dimensional latent spaces are particularly preferred due to their reduced computational demands for the calculation of the Riemannian metric and subsequent geodesics. However, higher dimensional latent spaces may also be of interest, as they have a larger capacity to encode trajectories with high-dimensional perceptual data (e.g. images). One question that arises in this context is what the learned manifold or metric is representing, given that the latent space may combine perceptual and motion data. Additionally, it is important to consider whether disentangled latent representations are necessary in such cases. Thus, the potential extension of our work involves learning a latent representation of task/joint space trajectories and exploring the integration of perceptual information into high-dimensional latent spaces.

The effective application of such extensions in higher dimensional latent spaces hinges on the careful selection of an appropriate real-time geodesic approximation and predictive variances modeling techniques. Our work has explored various approaches for approximating geodesics, including graph-based and gradient descent-based methods, which are potential candidates for achieving this objective. In particular, our gradient-based approach exhibits scalability to high-dimensional spaces, and existing literature provides evidence supporting the scalability of graph-based approaches up to $20$ dimensions~\citep{Chen2019:FastApproximateGeodesics}. Moreover, our model relies on RBF networks for modeling predictive variances, which are known to scale poorly to high-dimensional settings~\citep{Arvanitidis:LatentSO}. There is thus a need for both improving uncertainty estimation in generative models, as well as geodesic computations, in order to robustly scale to high-dimensional latent spaces.

\subsubsection*{Latent space dimensionality:}
This paper presents our use of a $2$-dimensional latent space in a task-space setting and a $3$-dimensional latent space in a joint-space setting. The choice of dimensionality for the latent space was based on empirical observations, and as such, may vary across different tasks and spaces. A more analytical approach may involve determining the minimum number of dimensions required to encode the data accurately. For instance, there could exist a correlation between a system's relative degree and the dimensionality of the latent space. Specifically, a system with a low relative degree is generally less nonlinear and may be represented with fewer dimensions, while a system with a high relative degree is more nonlinear and may require a higher-dimensional latent space to accurately capture its behavior. Nonetheless, the dimensionality of the latent space is not solely dependent on a system's relative degree, as other factors such as task complexity, required accuracy, and precision, amongst others, may also play a relevant role.

\subsubsection*{Choice of distributions:}
Using an isotropic Gaussian kernel, in which the covariance matrix is just a multiple of the identity matrix, is computationally simpler and more efficient than using an anisotropic Gaussian kernel. This is because the isotropic version only requires the specification of a single variance parameter, whereas an anisotropic Gaussian kernel needs the specification of a full covariance matrix, which is more computationally costly and requires more parameters to optimize. Additionally, the isotropic Gaussian kernel may be more interpretable, as it has a simpler functional form and only depends on a single variance parameter. However, the use of an isotropic Gaussian kernel might compromise the model's accuracy, as it does not allow for capturing correlations between different dimensions of the input space. Moreover, it is worth noting that the von Mises-Fischer distribution has been leveraged as an approximation for data distribution on the hypersphere when encoding quaternion data. However, it should be noted that there exist more expressive and accurate distributions, such as the Bingham distribution, which may provide more accurate modeling of the true distribution of the observed quaternion data. In the future, we plan to investigate the impact of using different distributions in the position, orientation, and joint spaces.